\documentclass[10pt,twocolumn,letterpaper]{article}

\usepackage[pagenumbers]{cvpr} 

\usepackage[accsupp]{axessibility}  

\usepackage{pifont}
\newcommand{\cmark}{\textcolor{green}{\ding{51}}}
\newcommand{\xmark}{\textcolor{red}{\ding{55}}}

\usepackage{color, colortbl}
\definecolor{Gray}{gray}{0.93}
\definecolor{Orange}{rgb}{1,0.5,0}
\definecolor{DGray}{gray}{0.83}
\definecolor{LightCyan}{rgb}{0.88,1,1}

\definecolor{WarnREd}{rgb}{1,0.4,0.4}
\definecolor{WarnOrange}{rgb}{1,0.682,0.502}
\definecolor{WarnPink}{rgb}{0.9176, 0.7215, 0.7215}
\definecolor{GoodGreen}{rgb}{0.5019, 0.9215, 0.6039}

\usepackage[T1]{fontenc}

\definecolor{styleblue}{HTML}{504099}
\definecolor{mypurple}{HTML}{9391ff}

\usepackage{algorithm}
\usepackage{algpseudocode}
\algrenewcommand\algorithmicrequire{\textbf{Input:}}
\algrenewcommand\algorithmicensure{\textbf{Return:}}

\usepackage{nth}
\usepackage{wrapfig}

\usepackage{multirow,mathtools }

\usepackage{pifont}
\usepackage{color, colortbl}

\usepackage{blindtext}
\usepackage{lipsum}

\usepackage{multirow}
\usepackage{listings}

\usepackage{bbm}

\usepackage [english]{babel}
\usepackage [autostyle, english = american]{csquotes}
\usepackage[nottoc]{tocbibind}

\usepackage{pifont}
\usepackage{url}
\usepackage[most]{tcolorbox}

\usepackage{lipsum}
\usepackage{soul}
\usepackage{xcolor}
\usepackage{wrapfig}
\usepackage{multirow,mathtools } 

\usepackage{adjustbox}
\MakeOuterQuote{"}

\usepackage{microtype}
\usepackage{graphicx}
\usepackage{booktabs} 

\usepackage{amsmath}

\definecolor{cvprblue}{rgb}{0.21,0.49,0.74}
\usepackage[pagebackref,breaklinks,colorlinks,allcolors=cvprblue]{hyperref}
\usepackage[capitalize,noabbrev]{cleveref}

\usepackage{tablefootnote}

\usepackage{threeparttable}
\usepackage[normalem]{ulem}

\usepackage{amsmath,amsfonts,bm}

\def\eqref#1{(\ref{#1})}

\def\1{\bm{1}}

\DeclareMathAlphabet{\mathsfit}{\encodingdefault}{\sfdefault}{m}{sl}
\SetMathAlphabet{\mathsfit}{bold}{\encodingdefault}{\sfdefault}{bx}{n}

\DeclareMathOperator*{\argmax}{arg\,max}

\newcommand{\bdelta}{\boldsymbol\delta}

\newcommand{\bx}{\mathbf{x}}

\newcommand{\bz}{\mathbf{z}}
\newcommand{\bw}{\mathbf{w}}
\newcommand{\ours}{\textsc{FaceLock}}
\newcommand{\cvl}{\textsc{CVLFace}}

\usepackage{circledsteps}
\usepackage{pifont}

\def\confName{CVPR}
\def\confYear{2025}

\title{Edit Away and My Face Will not Stay: \\Personal Biometric Defense against Malicious Generative Editing 
}

\author{
    Hanhui Wang$^{1, *}$, 
    Yihua Zhang$^{2, *}$, 
    Ruizheng Bai$^{3}$, 
    Yue Zhao$^{1}$, 
    Sijia Liu$^{2}$$^{\dagger}$, 
    Zhengzhong Tu$^{3}$$^{\dagger}$\\[0.2cm]
    $^1$University of Southern California \quad
    $^2$Michigan State University \quad
    $^3$Texas A\&M University \\[0.2cm]
   {\small\texttt{hanhuiwa@usc.edu, zhan1908@msu.edu, liusiji5@msu.edu, tzz@tamu.edu}}\\
    {\small $^*$ Equal contribution\quad $^{\dagger}$ Corresponding authors}
}

\begin{document}

\twocolumn[{%
\renewcommand\twocolumn[1][]{#1}%

\maketitle

\begin{center}
    \vspace*{-1.em}
    \includegraphics[width=0.94\textwidth]{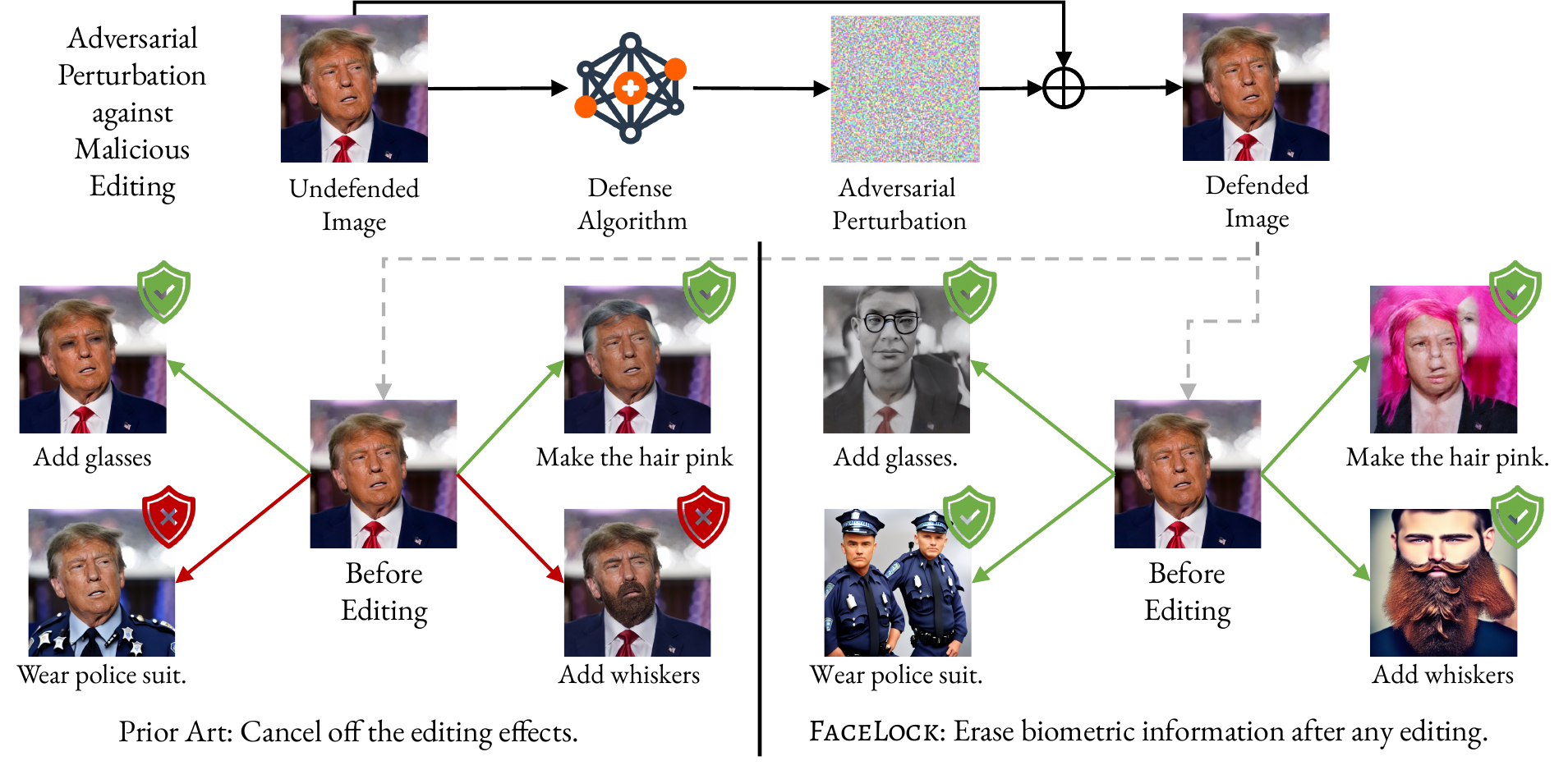}
    \vspace*{-1.em}
    \captionof{figure}{An illustration of adversarial perturbation generation for safeguarding personal images from malicious editing. Perturbations generated by prior work~\cite{salman2023raising, chen2023editshield} aim to cancel off editing effects, resulting in instability due to the diversity of editing instructions. In contrast, {\ours} does not prevent edits from being applied but instead erases critical biometric information (\textit{e.g.}, human facial features) after editing, making it agnostic to specific prompts and achieving superior performance.}
    \label{fig:teaser}
\end{center}
}]

\begin{abstract}
Recent advancements in diffusion models have made generative image editing more accessible than ever. While these developments allow users to generate creative edits with ease, they also raise significant ethical concerns, particularly regarding malicious edits to human portraits that threaten individuals' privacy and identity security. Existing general-purpose image protection methods primarily focus on generating adversarial perturbations to nullify edit effects. However, these approaches often exhibit instability to protect against diverse editing requests. In this work, we introduce a novel perspective to personal human portrait protection against malicious editing. Unlike traditional methods aiming to prevent edits from taking effect, our method, {\ours}, optimizes adversarial perturbations to ensure that original biometric information---such as facial features---is either destroyed or substantially altered post-editing, rendering the subject in the edited output biometrically unrecognizable. Our approach innovatively integrates facial recognition and visual perception factors into the perturbation optimization process, ensuring robust protection against a variety of editing attempts. Besides, we shed light on several critical issues with commonly used evaluation metrics in image editing and reveal cheating methods by which they can be easily manipulated, leading to deceptive assessments of protection. Through extensive experiments, we demonstrate that {\ours} significantly outperforms all baselines in defense performance against a wide range of malicious edits. Moreover, our method also exhibits strong robustness against purification techniques. Comprehensive ablation studies confirm the stability and broad applicability of our method across diverse diffusion-based editing algorithms. Our work not only advances the state-of-the-art in biometric defense but also sets the foundation for more secure and privacy-preserving practices in image editing. The code is publicly available at: \href{https://github.com/taco-group/FaceLock}{https://github.com/taco-group/FaceLock}.
\end{abstract}    
\section{Introduction}
\label{sec:intro}

Image editing has advanced at an unprecedented rate due to the rise of diffusion-based techniques, making it possible to produce edits that are indistinguishable from reality~\cite{kawar2023imagic, zhang2022sine, Zhang2023MagicBrush, hertz2022prompt, huang2024smartedit,fu2024mgie, brooks2023instructpix2pix, Choi2023CustomEditTI, li2024light, han2023highlypersonalizedtextembedding, ruiz2022dreambooth, gu2023photoswap, liu2023swapanything,qi2024spire}. This rapid development has led to tools capable of seamlessly modifying visual content, with edits so convincing that they are often impossible to differentiate from the original image. While this progress opens up creative possibilities, it also brings significant ethical and societal challenges.

The power of these editing techniques has led to severe ethical implications~\cite{zohny2023ethics, vyas2024ethical, Lawton2024,huang2025trustworthiness,xing2024autotrust}. Recent incidents, such as the widely discussed manipulation of Taylor Swift's images~\cite{nyt_taylor_swift_2024} and the proliferation of pornographic content affecting Korean schools~\cite{korean_deepfake_crisis_2024}, underscore the urgent need to address the risks associated with malicious image editing. These incidents have highlighted growing concerns about how personal images, particularly those depicting individuals' faces, can be misused once they are posted online~\cite{zhao2023diffswap, gu2023photoswap, liu2023swapanything}. Protecting such images from unauthorized and malicious edits has thus become an important topic of research~\cite{tang2019faces, an2024sd4privacy, he2024diff}.

To address this challenge, several recent attempts~\cite{salman2023raising,shan2023glaze, chen2023editshield, wang2023distraction, huang2023nightshade, liang2023adversarial, zhang2023robustness} have focused on using adversarial perturbations, which are imperceptible to human eyes but are intended to negate the effects of editing when such images are used as inputs to diffusion-based editing algorithms. These perturbations aim to protect personal images by preventing the success of the intended edits (see \textbf{Fig.\,\ref{fig:teaser}} for an illustration). However, current methods suffer from instability~\cite{salman2023raising, chen2023editshield, liang2023adversarial, zhang2023robustness} and simple purification methods. Specifically, while they are effective for certain types of editing instructions, they fail against others, largely due to the inherent diversity and versatility of editing prompts. The underlying issue is that as long as existing methods continue to focus on `canceling off editing effects', the inconsistency of results is inevitable. The diversity in editing prompts and the complexity of generative diffusion models make it difficult for such approaches to generalize effectively.

The rationale behind current defense methods is to ensure that the edited image does not meet the requirements of a successful image editing task. To understand this more deeply, we first revisit what constitutes a successful image editing task: it should accurately reflect the editing instruction while preserving the original, irrelevant visual features, such as those related to the subject's identity, including facial features. The latter requirement, which has been largely overlooked, provides an opportunity for a new defense strategy. Instead of attempting to cancel out edits, here, we ask: 

\begin{tcolorbox}[before skip=2mm, after skip=0.0cm, boxsep=0.0cm, middle=0.0cm, top=0.1cm, bottom=0.1cm]
    \textit{\textbf{(Q)}
    Can we design adversarial perturbations that cause edited images to lose their biometric information, making the edited image biometrically unrecognizable and thereby causing the edit to fail?
    }
\end{tcolorbox}
\vspace*{0.2cm}

Through a series of algorithmic designs, we demonstrate that creating adversarial perturbations that disrupt facial recognition while also introducing distinct visual disparities in facial features is far from trivial. To address this, we propose {\ours}, which strategically integrates a state-of-the-art facial recognition model into the diffusion loop as an adversary while also penalizing feature embeddings to achieve visual dissimilarity. By doing so, our method not only disrupts facial recognition but also ensures significant visual differences from the original, providing robust protection against malicious editing, see \textbf{Fig.\,\ref{fig:teaser}} for a comparison between {\ours} and prior arts. To this end, we summarize our contributions as follows:

\noindent $\bullet$ We present a novel perspective for protecting personal images from malicious editing, focusing on making biometric features unrecognizable after edits.

\noindent $\bullet$ We develop a new algorithm, {\ours}, that incorporates facial recognition models and feature embedding penalties to effectively protect against diffusion-based image editing.

\noindent $\bullet$ We conduct a critical analysis of the quantitative evaluation metrics commonly used in image editing tasks, exposing their vulnerabilities and highlighting the potential for manipulation to achieve deceptive results.

\noindent $\bullet$ Through extensive experiments, we demonstrate that {\ours} effectively alters human facial features against various editing prompts, achieving superior defense performance compared to baselines. We also show that {\ours} generalizes well to multiple diffusion-based algorithms and exhibits inherent robustness against purification methods.

\section{Related Work}
\label{sec:related}

\noindent \textbf{Generative editing models.} Recent advances in latent diffusion models ~\cite{rombach2022high} have demonstrated superior image editing capabilities through instructions and prompt editing ~\cite{kawar2023imagic, zhang2022sine, Zhang2023MagicBrush, hertz2022prompt}. Most recent methods ~\cite{huang2024smartedit,fu2024mgie} combine diffusion models with large language models for understanding text prompts. InstructPix2Pix ~\cite{brooks2023instructpix2pix} leverages a fine-tuned version of GPT-3 and images generated from SD and achieves on-the-fly image editing without further per-sample finetuning. On the other hand, many such models also allow personalized image editing ~\cite{Choi2023CustomEditTI, han2023highlypersonalizedtextembedding}. DreamBooth ~\cite{ruiz2022dreambooth} learns a unique identifier and class type of an object by finetuning a pretrained text-to-image model with a few images. SwapAnything ~\cite{liu2023swapanything} and Photoswap ~\cite{gu2023photoswap} allow for personal content editing by swapping faces and objects between two images. In the generative era, these tools offer unprecedented creative freedom but also raise ethical questions on privacy and malicious image editing, which motivate us to conduct this work.

\noindent \textbf{Defense against malicious editing.} Adversarial samples are clean samples manipulated intentionally to fool a machine learning model, often done by perturbing the image with an imperceptible small noise. Under a white-box setting, gradient-based methods, such as fast gradient sign method (FGSM), projected gradient decent (PGD)~\cite{nesterov2013introductory} and Carlini \& Wagner (CW) attack~\cite{carlini2017towards}, are among the most effective techniques in generating adversarial examples in classification models. Recent works like PhotoGuard, Editshield, AdvDM~\cite{salman2023raising, chen2023editshield, liang2023adversarial} have extended gradient-based methods to diffusion models and aim to protect images from malicious editing. PhotoGuard demonstrated an effective encoder attack mechanism by perturbing the source image towards an unrelated target image, e.g. an image of gray background. In particular, let $\mathcal{E}$ be the encoder, $\mathbf{z}_{\text{target}}$ be the latent representation of the target image. Under a attack budget $\epsilon$, PhotoGuard aims to optimize:
\vspace{-2pt}
\begin{align}
    \delta_{\text{Encoder}} = \arg\min_{\mathbf{\|\delta\|_{\infty}} \leq \epsilon} \|\mathcal{E}(\mathbf{x} + \delta) - \mathbf{z}_{\text{target}}\|.
\end{align}
Yet, the protection can be less effective if the image is slightly transformed. PhotoGuard takes a step further by considering expectations over transformation:
\vspace{-2pt}
\begin{align}
\max_{\mathbf{x_p}} \mathbb{E}_{f \sim \mathcal{F}} \left[ \text{Dist}(\mathcal{E}(f(\mathbf{x_p})), \mathcal{E}(\mathbf{x})) \right] - \beta \cdot \| \mathbf{x_p} - \mathbf{x} \|_2^2,
\end{align}
where $\mathbf{x}$ is the source image, $\mathbf{x_p}$ is the perturbed image, and $\mathcal{F}$ is a distribution over a set of transformations. 

However, these approaches are typically less robust, as the gradients are highly dependent on model architecture and parameters. Distraction Is All you Need~\cite{wang2023distraction} circumvent this by attacking the cross attention mechanism between image and editing instruction, so diffusion models misinterpret the target editing regions. Glaze and Nightshade~\cite{shan2023glaze, huang2023nightshade} instead perturb the image towards a completely different image with another style or concept. These approaches make the image less susceptible to the specificities of model architecture and is generally more robust across different models.

The adversarial techniques mentioned above primarily protect portrait images by interfering with the editing process. However, nullifying the editing process does not always safeguard facial features or biometric information. We propose a novel way of protecting images by incorporating facial recognition model into the perturbation process. Although the model is capable of editing the image according to the prompt, we ensure that the facial features are altered or destroyed during the process.

\noindent \textbf{Facial recognition.} Recent facial recognition works~\cite{kim2022adaface, deng2019arcface, wang2018cosface, Boutros_2022_CVPR, huang2020curricularface, QMagFace, meng2021magface} have proposed several margin-based softmax loss functions to enhance the discriminative power and feature extraction ability of facial recognition models. \cvl~\cite{kim2022adaface} utilizes these models to extract features from two images and computes the cosine similarity between these features to verify a person's identity. In addition, recent works ~\cite{Sface_Boutros,10454585,BoutrosKFKD23,10208805, qiu2021synface} also leverage synthetic images during training for enhanced privacy protection, highlighting the need for privacy protection in image editing as well. Our approach builds on top of CVLFace and protects biometric information by minimizing the cosine similarity between features. Our work is the first application of facial recognition on perturbation generation, enabling a new axis of identity protection.
\section{{\ours}: Adversarial Perturbations for Biometrics Erasure}
\label{sec: method}

\begin{figure}[htb]
    \centering
    \vspace*{-1.5em}
    \includegraphics[width=\linewidth]{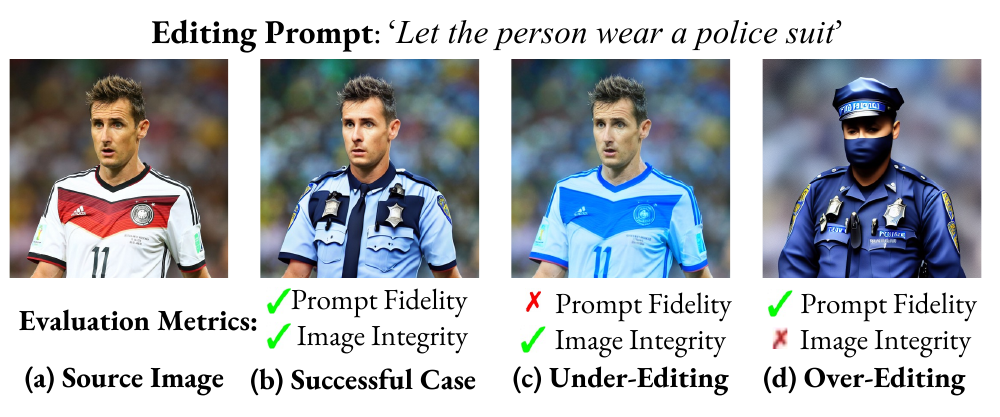}
    \vspace*{-2.em}
    \caption{Illustration of the two requirements of image editing task: prompt fidelity and image integrity. (a) Source image before editing; (b) A successful editing example holding both metrics; (c) Failure case due to the lack of prompt fidelity leading to under-editing and (d) the lack of the image integrity leading to over-editing.
    }
    \vspace*{-1em}
    \label{fig: edit_criteria}
\end{figure}

\noindent \textbf{What defines a successful image editing task?} Before introducing our proposed method for safeguarding human portrait images from malicious edits, we revisit the criteria for a successful image editing outcome. Specifically, we propose that a successful text-guided image editing hinges on two critical requirements: \ding{182} \textbf{prompt fidelity}, and \ding{183} \textbf{image integrity}. \textbf{Prompt fidelity} requires that the edit accurately reflects the instructions provided in the prompt. For instance, as shown in \textbf{Fig.\,\ref{fig: edit_criteria}}, a successful edit replaces the person's clothing with a police uniform as instructed by the prompt. Meanwhile, \textbf{image integrity} requires that other elements in the image remain intact after editing. Although this requirement is less explicit than prompt fidelity, it defines the essence of image editing and differentiates it from general text-to-image generation tasks. As illustrated in Fig.\,\ref{fig: edit_criteria}, aside from the change in attire, the edited image should retain as much of the subject's original appearance as possible, including facial features, poses, and other details. While prompt fidelity has been emphasized and extensively studied~\cite{rombach2022high, ruiz2022dreambooth, brooks2023instructpix2pix, huang2024smartedit}, image integrity remains long-overlooked and underexplored in literature. Next, we will demonstrate how this holistic view of image editing can provide new insights into protecting human portraits from malicious edits.

\begin{figure*}[tbh]
    \centering
    \begin{tabular}{cccccc}
         \hspace*{-2mm} \includegraphics[width=0.14\textwidth]{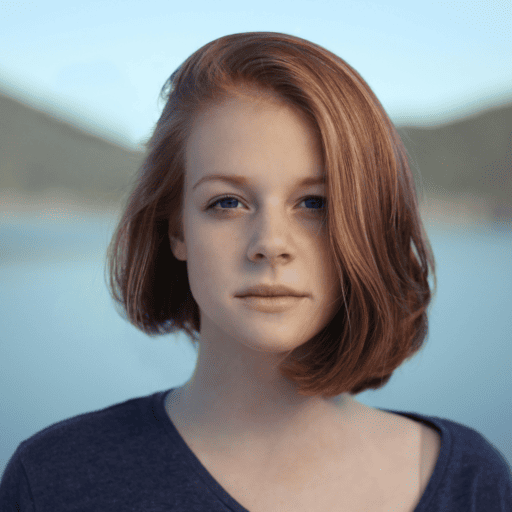} 
         & \hspace*{-2mm} \includegraphics[width=0.14\textwidth]{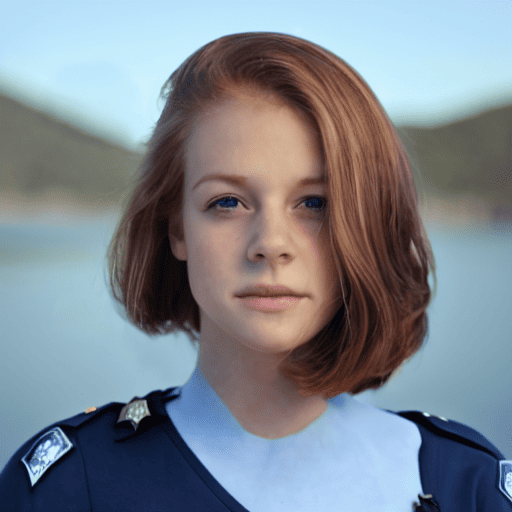} 
         & \hspace*{-2mm}  \includegraphics[width=0.14\textwidth]{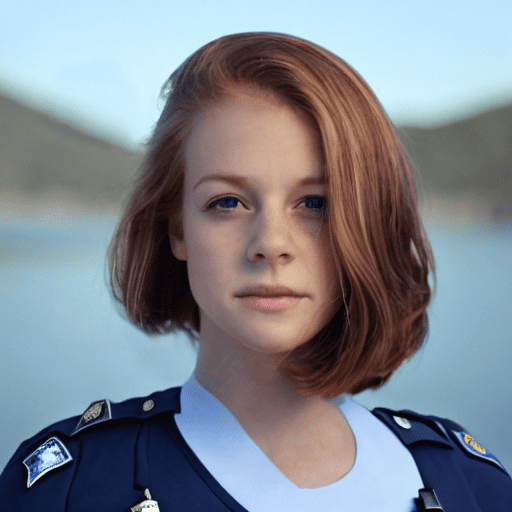} 
         & \hspace*{-2mm}  \includegraphics[width=0.14\textwidth]{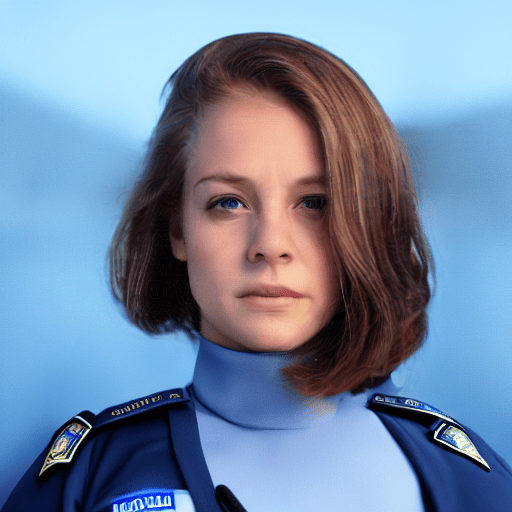} 
         & \hspace*{-2mm}  \includegraphics[width=0.14\textwidth]{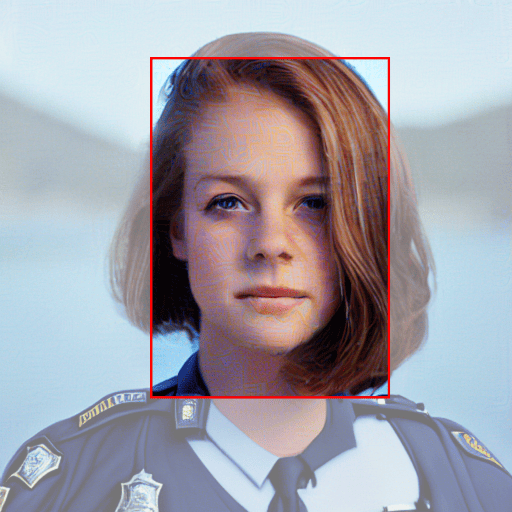} 
         & \hspace*{-2mm}  \includegraphics[width=0.14\textwidth]{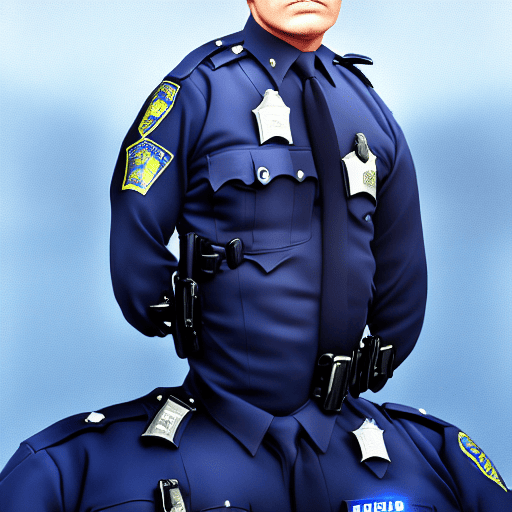}\\
         \hspace*{-2mm} \footnotesize{FR=1.0}
         & \hspace*{-2mm}  \footnotesize{FR=\textcolor{red}{0.972}}
         & \hspace*{-2mm} \footnotesize{FR=\textcolor{red}{0.901}}
         & \hspace*{-2mm}  \footnotesize{FR=0.273}
         & \hspace*{-2mm}  \footnotesize{FR=\textcolor{red}{0.658}}
         & \hspace*{-2mm}  \footnotesize{FR=0.093} \\
         \hspace*{-2mm} 
         & \hspace*{-2mm}  \footnotesize{{\xmark} Visual Change}
         & \hspace*{-2mm}  \footnotesize{{\xmark} Visual Change}
         & \hspace*{-2mm}  \footnotesize{{\xmark} Visual Change}
         & \hspace*{-2mm}  \footnotesize{{\xmark} Visual Change}
         & \hspace*{-2mm}  \footnotesize{{\cmark} Visual Change} \\
         \hspace*{-2mm} \footnotesize{(a) Source Image}
         & \hspace*{-2mm}  \footnotesize{(b) No Protection}
         & \hspace*{-2mm}  \footnotesize{(c) Design I: CVL}
         & \hspace*{-2mm}  \footnotesize{(d) Design II: CVL-D}
         & \hspace*{-2mm}  \footnotesize{(e) Design III: CVL-D + Pixel}
         & \hspace*{-2mm}  \footnotesize{(f) {\ours}}
    \end{tabular}
    \vspace*{-0.7em}
   \caption{Source and edited images generated from different protection methods based on the instruction \textquotedblleft\textit{Let the person wear a police suit}\textquotedblright. The FR score below each image represents the facial representation similarity between the edited and source images and scores marked in \textcolor{red}{red} indicate insignificant changes biometric recognition results by {\cvl} compared to source image. `CVL' refers to perturbations generated targeting the {\cvl} model alone. `CVL-D' represents protection targeting both the {\cvl} model and the diffusion model, while `CVL-DP' incorporates an auxiliary loss to enforce pixel-level disparity between the edited and source images. {\ours} targets the {\cvl} and diffusion model, aiming to enhance the disparity between the feature embeddings of the decoded and original images.}
    \label{fig: method_development}
    \vspace*{-1.2em}
\end{figure*}

\noindent \textbf{A new direction for defending against malicious editing.} As discussed above, to safeguard personal images from malicious editing, the defender must ensure that at least one of the two requirements is not met. Previous works have primarily focused on generating adversarial perturbations to prevent edits from taking effect, thereby reducing prompt fidelity~\cite{salman2023raising, chen2023editshield, shan2023glaze, liang2023adversarial, huang2023nightshade}. However, these approaches often suffer from instability and are effective only for a limited range of editing instructions, resulting in poor generalization. The core issue is the versatility of editing instructions—making it unlikely that a single perturbation can defend against all potential prompts. Therefore, we explore a new direction: optimizing perturbations to destroy biometric information after editing, rendering the edited image biometrically unrecognizable and thereby causing the edit to fail.

\noindent \textbf{Adversarial perturbation for facial disruption is nontrivial.} The goal of our defense method is to disrupt human facial features during the sampling process in diffusion-based editing models. \textbf{Design I (CVL): Perturbation against facial recognition models.} A straightforward approach is to apply an adversarial perturbation against a state-of-the-art (SOTA) facial recognition model, such as the {\cvl} model~\cite{kim2022adaface}, and use the perturbed image as input to the image editing model. However, as shown in \textbf{Fig.\,\ref{fig: method_development}(c)}, the perturbation that successfully fools the {\cvl} model does not persist through the diffusion model's sampling process, resulting in an edited image with minimal disruption to facial features, as indicated by both the high facial similarity (FR) score and the visually similar appearance. The underlying issue with this approach is that the perturbations are generated independently of the diffusion process. Prior work~\cite{nie2022diffusion} highlights that diffusion models possess an inherent ability to "purify" adversarial perturbations through their sampling process. 

\noindent \textbf{Design II (CVL-D): Perturbation against diffusion with {\cvl} model in the loop.} To address this, we incorporate the {\cvl} model into the diffusion process and design a method to directly interfere with the sampling stage. Given the high computational costs of disrupting the entire diffusion process, we instead bypass this step once the latent representation of the input image is obtained. The perturbation is then optimized by a facial recognition loss that maximizes the biometric disparity between the decoded image and the source input:

\vspace*{-1em}
{\small
\begin{align}
    \bdelta = \argmax_{\|\bdelta\|_\infty \leq \epsilon} f_\text{FR}(\mathcal{D}(\mathcal{E}(\mathbf{x} + \bdelta)), 
    \mathbf{x}),
\end{align}
}

\noindent where $\mathcal{D}$ and $\mathcal{E}$ denote the decoder and encoder used by the diffusion model, respectively, and $f_\text{FR}(\cdot, \cdot)$ computes the facial similarity score. As shown in \textbf{Fig.\,\ref{fig: method_development}(d)}, while this method significantly reduces the facial recognition similarity score, the edited image still resembles the original subject, suggesting room for further improvement in visual effects. 

\noindent \textbf{Design III (CVL-DP): Perturbation against diffusion facial similarity with pixel-level penalty.} To enhance the visual disparity between the edited image and the source image, we introduce a pixel-level loss focused on facial regions defined by a mask:

\vspace*{-1em} {\small \begin{align} \bdelta = \argmax_{\|\bdelta\|_\infty \leq \epsilon} f_\text{FR}(\mathcal{D}(\mathcal{E}(\mathbf{x} + \bdelta)),\mathbf{x}) + \lambda \|\bdelta \odot \mathbf{m}\|_2, \end{align} \vspace*{-1em} }

\noindent where $\mathbf{m}$ defines the facial region extracted by the {\cvl} model. However, as shown in \textbf{Fig.\,\ref{fig: method_development}(e)}, the pixel-level loss primarily results in color shifts rather than significant distortion of the subject's facial features. This limitation motivated the development of {\ours}, which aims to generate perturbations that enhance both facial dissimilarity scores and visual facial discrepancies.

\noindent\textbf{{\ours}: Perturbation optimization on facial disruption and feature embedding disparity.} The lesson from CVL-DP indicates that pixel-level changes do not necessarily lead to distinct visual facial features. Thus, we transition to a more effective feature-level approach, using pretrained convolutional neural networks to extract and maximize the difference between high-level feature embeddings of the decoded and source image:

\vspace*{-1em} {\small \begin{align} \hspace*{-1em}\bdelta = \argmax_{\|\bdelta\|_\infty \leq \epsilon} f_\text{FR}(\mathcal{D}(\mathcal{E}(\mathbf{x} + \bdelta)),\mathbf{x}) + \lambda f_\text{FE}(\mathcal{D}(\mathcal{E}(\mathbf{x} + \bdelta), \mathbf{x}), 
\label{eq: facelock}
\end{align} 
\vspace*{-1em}}

\noindent where $f_\text{FE}(\cdot, \cdot)$ extracts feature embeddings from the input images and compute the distance between them. To solve \eqref{eq: facelock}, the widely used projected gradient descent (PGD)~\cite{nesterov2013introductory} method can be employed. We refer more implementation details in Sec.\,\ref{sec: experiments}.
\section{Pitfalls in The Widely-Used Quantitative Evaluation Metrics for Image Editing Tasks}
\label{sec:evaluation}

In this section, we begin by providing a critical analysis of existing quantitative evaluation metrics for image editing tasks~\cite{wang2004image,radford2021learning,zhang2024unlearncanvas,tu2021ugc}. For the first time, we highlight potential pitfalls in these widely accepted metrics, particularly how they can be easily manipulated to achieve deceptively high scores. Finally, we introduce two new, more robust metrics for evaluating human portrait editing. Detailed mathematical descriptions of the quantitative evaluation metrics discussed in this section can be found in Appx.~\ref{app: exp_setup}.

\noindent \textbf{Existing quantitative metrics suffer from pitfalls and can be manipulated for misleading performance.} As discussed in \S\ref{sec: method}, the evaluation of general image editing tasks should consider two aspects: prompt fidelity and image integrity. However, all existing quantitative metrics, including CLIP scores~\cite{radford2021learning}, SSIM, and PSNR primarily focus on the former, namely how well the editing instruction is reflected in the edited image. In the following, we revisit each of these metrics and demonstrate the intrinsic pitfalls in their design.

\begin{table*}[htb]
  \centering
  \small
      \renewcommand{\arraystretch}{0.9} 

  \caption{Quantitative evaluation on prompt fidelity (CLIP-S, PSNR, SSIM, LPIPS) and image integrity (CLIP-I, FR). Arrows ($\uparrow$ or $\downarrow$) indicate whether a higher or lower value is preferred for a successful defense. All results are averaged over 5 different random seeds for editing. Results in the form $a$\footnotesize{$\pm b$} \small represent mean $a$ with std $b$. The best result within each evaluation metric is highlighted in \textbf{bold}.}  \label{tab:main}
  \vspace{-0.8em}
  \begin{tabular}{@{}lcccccc@{}}
    \toprule
    & \multicolumn{4}{c}{\textbf{Prompt Fidelity}} & \multicolumn{2}{c}{\textbf{Image Integrity}}\\
    \cmidrule(lr){2-5} \cmidrule(lr){6-7}
    \textbf{Method} & $\textbf{CLIP-S}\downarrow$ & $\textbf{PSNR}\downarrow$ & $\textbf{SSIM}\downarrow$ & $\textbf{LPIPS}\uparrow$ & $\textbf{CLIP-I}\downarrow$ & $\textbf{FR}\downarrow$\\
    \midrule
    \textbf{No Defense} & 0.118\footnotesize{$\pm$0.037} & - & - & - & 0.808\footnotesize{$\pm$0.074} & 0.833\footnotesize{$\pm$0.111} \\
    \midrule
    \textbf{PhotoGuard Encoder attack} & \textbf{0.108}\footnotesize{$\pm$0.030} & \textbf{15.44}\footnotesize{$\pm$2.01} & 0.612\footnotesize{$\pm$0.056} & 0.403\footnotesize{$\pm$0.071} & 0.670\footnotesize{$\pm$0.118} & 0.590\footnotesize{$\pm$0.264} \\
    \textbf{EditShield} & 0.110\footnotesize{$\pm$0.026} & 17.74\footnotesize{$\pm$2.20} & 0.593\footnotesize{$\pm$0.072} & 0.382\footnotesize{$\pm$0.071} & 0.677\footnotesize{$\pm$0.096} & 0.641\footnotesize{$\pm$0.231}\\
    \textbf{Untargeted Encoder attack} & 0.116\footnotesize{$\pm$0.023} & 16.74\footnotesize{$\pm$2.27} & \textbf{0.589}\footnotesize{$\pm$0.084} & 0.371\footnotesize{$\pm$0.094} & 0.653\footnotesize{$\pm$0.090} & 0.563\footnotesize{$\pm$0.236}\\
    \textbf{CW L2 attack} & 0.115\footnotesize{$\pm$0.031} & 19.64\footnotesize{$\pm$2.46} & 0.701\footnotesize{$\pm$0.060} & 0.247\footnotesize{$\pm$0.062} & 0.733\footnotesize{$\pm$0.089} & 0.725\footnotesize{$\pm$0.173}\\
    \textbf{VAE attack} & 0.114\footnotesize{$\pm$0.034} & 19.40\footnotesize{$\pm$1.70} & 0.715\footnotesize{$\pm$0.039} & 0.251\footnotesize{$\pm$0.060} & 0.786\footnotesize{$\pm$0.061} & 0.846\footnotesize{$\pm$0.097}\\
    \midrule
    \textbf{\ours~(ours)} & 0.114\footnotesize{$\pm$0.024} & 17.11\footnotesize{$\pm$2.36} & \textbf{0.589}\footnotesize{$\pm$0.079} & \textbf{0.436}\footnotesize{$\pm$0.065} & \textbf{0.648}\footnotesize{$\pm$0.089} & \textbf{0.315}\footnotesize{$\pm$0.109}\\
    \bottomrule
  \end{tabular}
  \vspace*{-1.5em}
\end{table*}

\begin{figure}[htb]
    \centering
    \begin{tabular}{@{}cccc@{}}
        \multicolumn{4}{c}{\footnotesize{\textbf{Editing Prompt}: `\textit{Let the person's hair turn pink}'.}}\\
        \hspace*{-2mm}\includegraphics[width=0.23\linewidth]{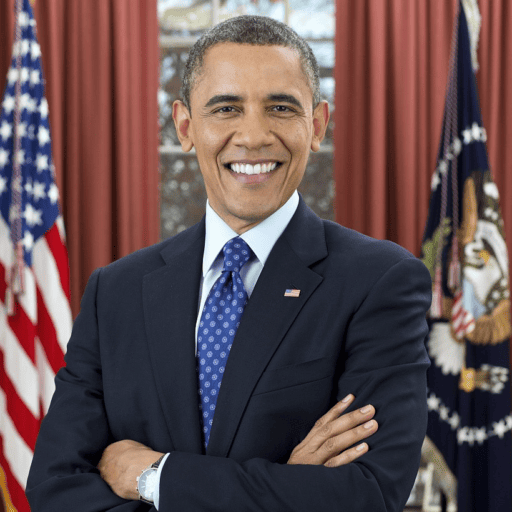}
        & \hspace*{-2mm}\includegraphics[width=0.23\linewidth]{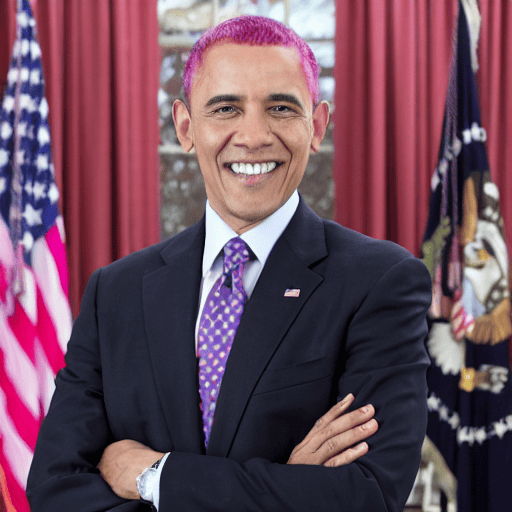}
        & \hspace*{-2mm}\includegraphics[width=0.23\linewidth]{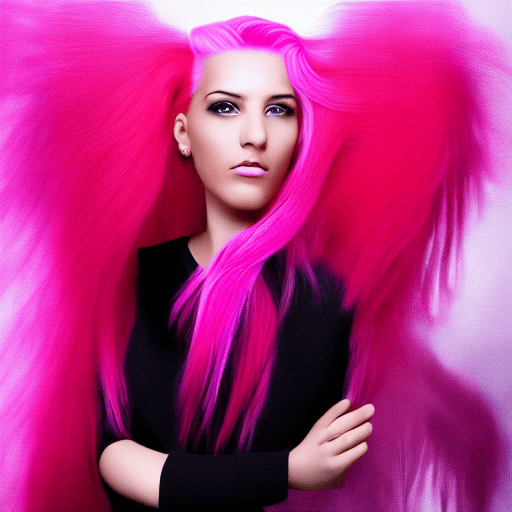}
        & \hspace*{-2mm}\includegraphics[width=0.23\linewidth]{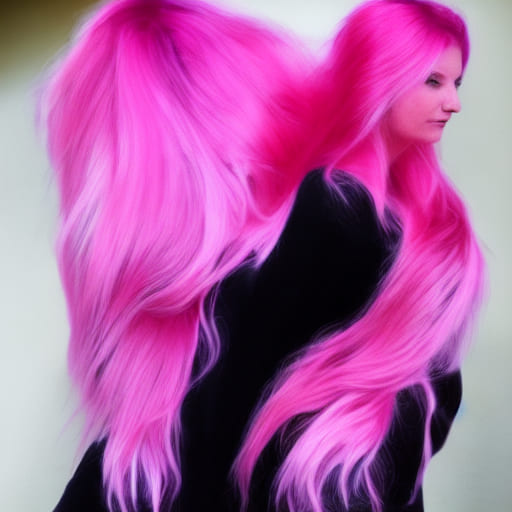}\\
        \hspace*{-2mm}\footnotesize{CLIP-S=N/A}
        & \hspace*{-2mm}\footnotesize{CLIP-S=0.091}
        & \hspace*{-2mm}\footnotesize{CLIP-S=0.103} 
        & \hspace*{-2mm}\footnotesize{CLIP-S=0.118} \\
        \hspace*{-2mm}\footnotesize{\textbf{(a) Source Image}}
        & \hspace*{-2mm}\footnotesize{\textbf{(b) Edited I}}
        & \hspace*{-2mm}\footnotesize{\textbf{(c) Edited II}}
        & \hspace*{-2mm}\footnotesize{\textbf{(d) Edited III}}\\

    \end{tabular}
    \vspace*{-1em}
    \caption{CLIP score (CLIP-S) of different editing results. The CLIP score provides a contradictory ranking (III > II > I) compared to the visual quality (I > II > III), as it overemphasizes the presence of elements from the editing prompt, thereby favoring over-editing.}
    \label{fig: clip_d}
    \vspace*{-1em}
\end{figure}

\noindent \textbf{CLIP-based scores overemphasize the presence of elements from the editing instructions, often prioritizing over-editing.} CLIP-based scores are widely used to assess prompt fidelity by measuring the cosine similarity between the CLIP text embedding of the editing prompt and the visual embedding difference between the edited and source images. While this metric effectively indicates whether the edit has taken effect, it tends to overemphasize the presence of specific elements in the edited image. \textbf{Fig.\,\ref{fig: clip_d}} shows a contradictory CLIP score ranking compared to the visual editing quality. Although Fig.\,\ref{fig: clip_d}(b) demonstrates a visually balanced outcome between the editing effect `turn the hair pink' and preserving other irrelevant (especially facial) features, the CLIP-based score still assigns higher values to Fig.\,\ref{fig: clip_d}(c) and (d) simply because they show stronger `pink hair' effects, even if the subject's identity has been completely altered. Therefore, CLIP-based scores can easily prioritize over-editing and be manipulated by replicating elements from the editing instructions.

\begin{figure}[htb]
    \vspace*{-1em}
    \centering
    \begin{tabular}{@{}cccc@{}}
        \multicolumn{4}{c}{\footnotesize{\textbf{Editing Prompt}: `\textit{Let the person wear a hat}'.}}\\
        \hspace*{-2mm}\includegraphics[width=0.23\linewidth]{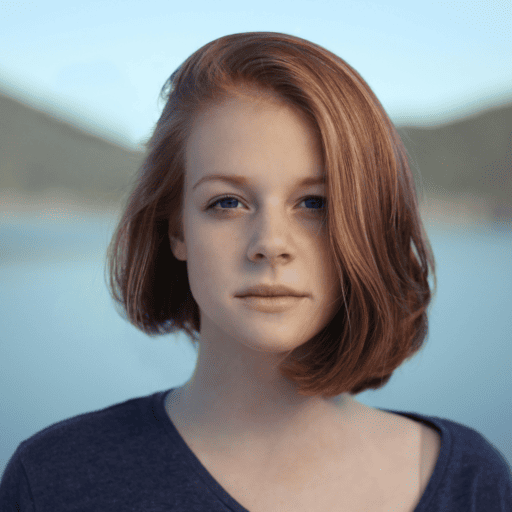}
        & \hspace*{-2mm}\includegraphics[width=0.23\linewidth]{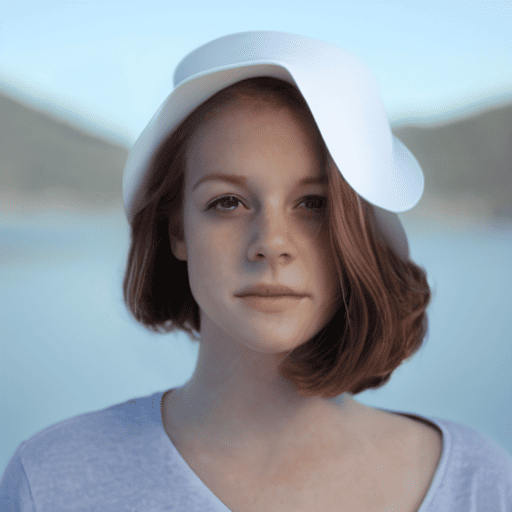}
        & \hspace*{-2mm}\includegraphics[width=0.23\linewidth]{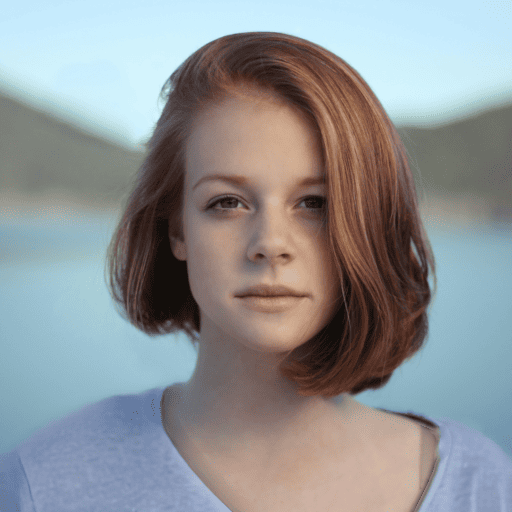}
        & \hspace*{-2mm}\includegraphics[width=0.23\linewidth]{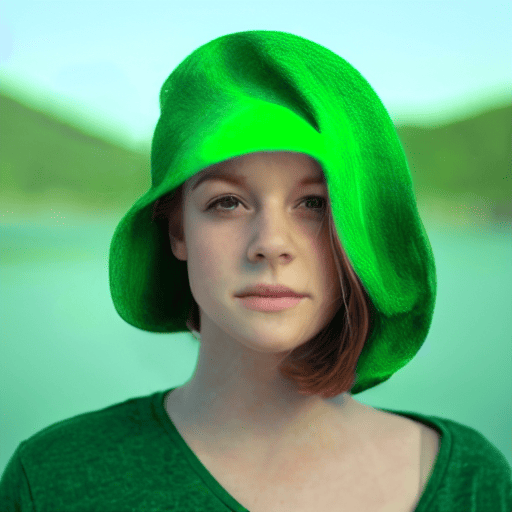}\\
        \multicolumn{2}{c}{\hspace*{-2mm}\footnotesize{SSIM=N/A}}
        & \hspace*{-2mm}\footnotesize{SSIM=0.869} 
        & \hspace*{-2mm}\footnotesize{SSIM=0.746} \\
        \multicolumn{2}{c}{\hspace*{-2mm}\footnotesize{PSNR=N/A}}
        & \hspace*{-2mm}\footnotesize{PSNR=16.44} 
        & \hspace*{-2mm}\footnotesize{PSNR=11.60} \\
        \hspace*{-2mm}\footnotesize{\textbf{(a) Source Image}}
        & \hspace*{-2mm}\footnotesize{\textbf{(b) No Defense}}
        & \hspace*{-2mm}\footnotesize{\textbf{(c) Defense I}}
        & \hspace*{-2mm}\footnotesize{\textbf{(d) Defense II}}
    \end{tabular}
    \vspace*{-1em}
    \caption{SSIM and PSNR scores of different defense methods. Although Defense I (b) demonstrates a successful defense, Defense II (d) is assigned a much lower (better) SSIM and PSNR score simply due to its larger pixel-level statistical difference from (b). SSIM and PSNR treat the edited image \textit{w/o} defense as the gold standard, without accounting for the diversity of possible editing outcomes, which can lead to a false sense of defense success.}
    \vspace*{-1em}
    \label{fig: psnr_ssim}
\end{figure}

\noindent \textbf{SSIM and PSNR over-rely on differences between the edited image and the undefended source, potentially leading to a false sense of successful defense.} Unlike CLIP-based scores, metrics such as SSIM and PSNR evaluate whether a defense against editing is successful by comparing the pixel-level statistical differences between the edited images with and without defense. While comparing against the edited image without defense can be effective in some scenarios, concluding that a defense is successful simply because the defended image differs from the undefended one is premature. 
For example, in \textbf{Fig.\,\ref{fig: psnr_ssim}}, Fig.\,\ref{fig: psnr_ssim}(b) demonstrates a successful edit based on the instruction `Let the person wear a hat.' While Fig.\,\ref{fig: psnr_ssim}(c) shows a genuinely successful defense, Fig.\,\ref{fig: psnr_ssim}(d) is incorrectly assigned a lower SSIM/PSNR score (where lower scores indicate better defense). This suggests a greater pixel-level statistical distance from Fig.\,\ref{fig: psnr_ssim}(b) compared to Fig.\,\ref{fig: psnr_ssim}(c). However, this assessment is flawed, as Fig.\,\ref{fig: psnr_ssim}(d) clearly represents a failed defense, given that a green hat has been applied to the source image. The pixel statistics-based score is misleading simply because the color of the hat differs from that in Fig.\,\ref{fig: psnr_ssim}(b). Therefore, treating the edited image without defense as a gold standard is risky, as the variability of editing effects, even with a single instruction, must be considered.

\noindent \textbf{LPIPS score as a more robust metric for prompt fidelity evaluation.} To address the limitations of pixel-level statistics used by SSIM and PSNR, we propose using the Learned Perceptual Image Patch Similarity (LPIPS~\cite{zhang2018unreasonable}) score to evaluate the similarity between edited images. Unlike traditional similarity metrics, LPIPS leverages pretrained neural networks to quantify perceptual differences by comparing high-level semantic features of images, offering a more robust assessment of protection effectiveness. We believe this approach can help mitigate the generalization issues associated with relying on a single reference image, as highlighted in the analysis above.

\noindent \textbf{Facial recognition similarity score for image integrity evaluation.} In this work, we propose evaluating image integrity as a means of assessing defense performance. However, we acknowledge that developing a cost-effective metric for general image integrity—defined as retaining all elements irrelevant to the editing—is challenging due to the diversity of elements present in an image. Therefore, we focus specifically on how well \textit{human facial details} are preserved after editing, under the assumption that facial features are not altered. For this purpose, we use the facial recognition (\textbf{FR}) similarity score to compare the subjects in the edited and source images. Generally, if the edited image does not statistically (in terms of FR score) and visually resemble the original subject, it indicates a successful defense. Additionally, we use the cosine similarity of the CLIP score (CLIP-I) between the edited and source images as a reference indicator on the general preservation effect.
\section{Experiments}
\label{sec: experiments}

\begin{figure*}[t]
    \centering
    \includegraphics[width=\linewidth]{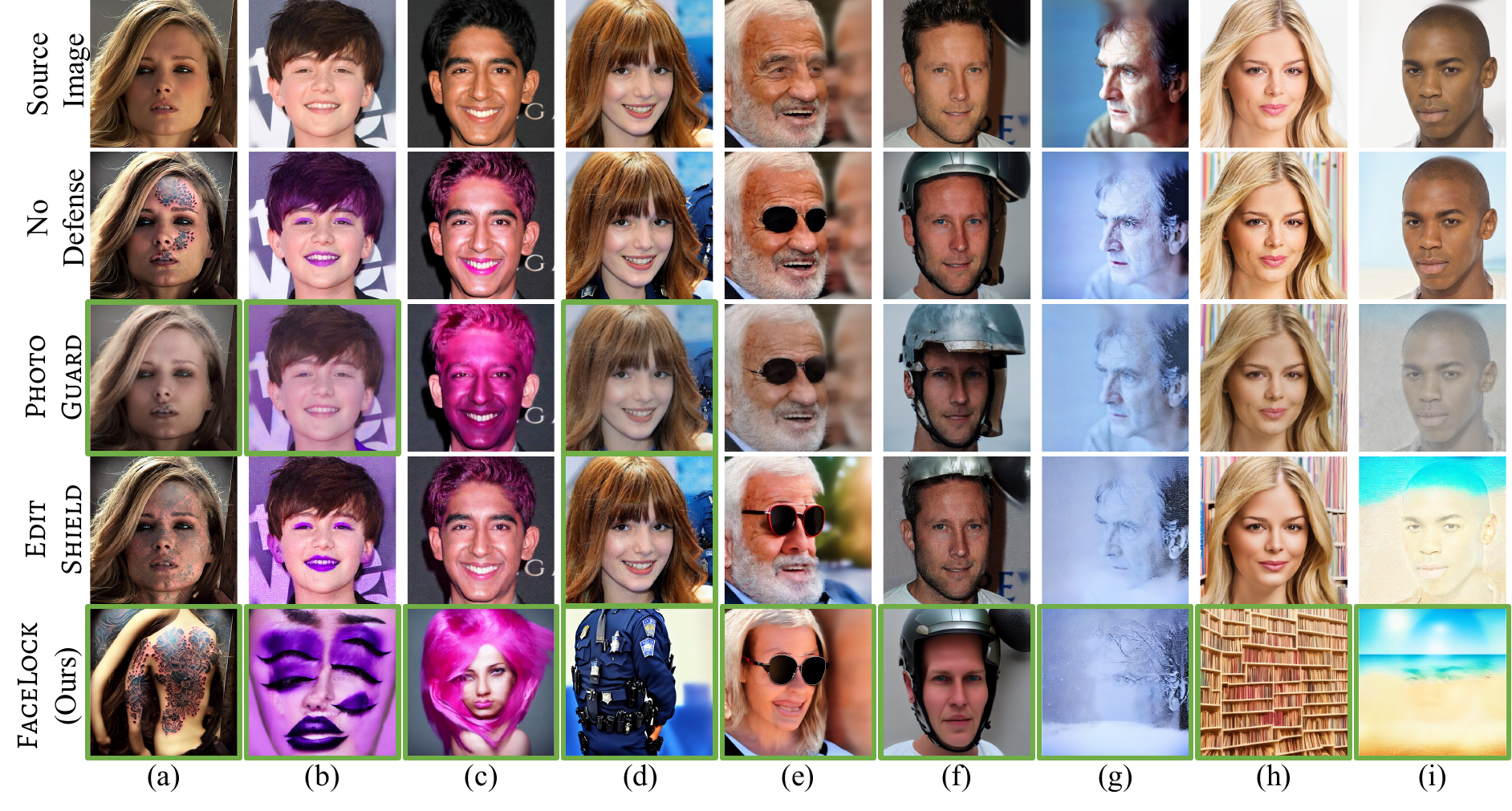}
    \vspace*{-2.5em}
    \caption{Qualitative results of different defense methods. Three editing types are included: \textbf{facial feature modifications} ((a) `Let the person have a tattoo'; (b) `Let the person wear purple makeup'; (c) `Turn the person's hair pink'), \textbf{accessory adjustments} ((d) `Let the person wear a police suit'; (e) `Let the person wear sunglasses'; (f) `Let the person wear a helmet'), and \textbf{background alterations} ((g) `Let it be snowy'; (h) `Set the background in a library'; (i) `Change the background to a beach'). Images in green frames denote successful defenses.}
    \vspace*{-1.5em}
    \label{fig:qualitative}
\end{figure*}

\subsection{Experiment Setup}

\noindent\textbf{Models and dataset.} We adopt the widely accepted InstructPix2Pix~\cite{brooks2023instructpix2pix} as our primary target model for prompt-based image editing. In our experiments, we utilize a filtered subset of the CelebA-HQ dataset~\cite{karras2017progressive}, a high-quality human face attribute dataset widely used in the facial analysis community. The dataset consists of $2,000$ human portrait images spanning diverse race, age, and gender groups. For editing prompts, we manually selected $25$ prompts across three categories: facial feature modifications (\textit{e.g.}, hair, nose modification), accessory adjustments (\textit{e.g.}, clothing, eyewear), and background alterations.

\noindent\textbf{Baselines.} We evaluate {\ours} against two established text-guided image editing protection methods: PhotoGuard \cite{salman2023raising} and EditShield \cite{chen2023editshield}, both designed for general image protection. Additionally, we also compare against a variety of widely used methods \cite{zhang2022revisiting,zhang2022fairness,zhang2022distributed,zhang2023robust,zhang2023generate,zhang2024defensive,zhuang2023pilot} in adversarial machine learning field, including untargeted encoder attack, CW attack, and VAE attack as other baseline methods. Full details on these baselines are provided in Appx.\,\ref{app: exp_setup}.

\noindent\textbf{Evaluation metrics.} We adopt quantitative evaluation metrics across two categories: prompt fidelity and image integrity. For prompt fidelity, we report PSNR, SSIM, and LPIPS scores between edits on protected and unprotected images, as well as the CLIP similarity score (CLIP-S), which captures the alignment between the edit-source image embedding shift and the text embedding. For image integrity, we report the CLIP image similarity score (CLIP-I) and facial recognition similarity score (FR). CLIP-I captures overall visual similarity, while FR specifically measures similarity in biometric information.

\noindent\textbf{Implementation details} 
For a fair comparison, we set the perturbation budget to 0.02 and the number of iterations to 100 for all methods, except EditShield, which does not have a default perturbation budget. Additionally, we include the untargeted latent-wise loss from EditShield as a regularization term to stabilize the protection results. Further experimental details are provided in Appx.\,\ref{app: exp_setup}.

\subsection{Experiment Results}

\noindent\textbf{Superior performance of \ours~in human portrait image protection: quantitative and qualitative evaluation.}
Building upon our analysis of comprehensive evaluation metrics for image editing and protection, we present a quantitative evaluation of various protection methods in \textbf{Tab.~\ref{tab:main}}. Our proposed method, \ours, demonstrates remarkable protection effectiveness across both prompt fidelity and image integrity metrics. Regarding prompt fidelity, \ours~achieves competitive results in multiple metrics. It ties the lowest SSIM score and maintains a competitive CLIP-S score, and more notably, it excels in the LPIPS metric with the highest score. This aligns with our discussion on the importance of perceptual measures over pixel-based metrics. For image integrity, \ours~outperforms all baselines significantly, especially in FR scores. This underscores its unparalleled efficacy in protecting the subject's biometric information against malicious editing. In \textbf{Fig.~\ref {fig:qualitative}}, we present qualitative results of the three editing types. As we can see, Our approach demonstrates the most pronounced alteration of biometric details between the edited and source images. For example, in the \textquotedblleft\textit{Let the person wear sunglasses}\textquotedblright\ editing scenario, while the edited image presents a person wearing sunglasses, it also transforms the individual from an elderly man in the source image to a young woman. These results indicate that {\ours} effectively protects images from different editing instructions.

\noindent \textbf{\ours~ demonstrates consistent protection across diverse editing types.} In \textbf{Tab.~\ref{tab:edit-type}}, we present the facial recognition similarity scores for three editing types: facial feature modifications, accessory adjustments, and background alterations. As shown, {\ours} provides robust protection across all three editing types. Notably, background alterations generally yield the highest facial recognition similarity scores across all methods and the clean edit scenario, indicating that just modifying the background is prone to preserve more facial identity compared to direct facial modifications. 
Despite the inherent challenge of protecting identity during background alterations, our method still achieves promising protection results in this category, demonstrating its effectiveness even in the most demanding scenarios.

\begin{table}[t]
    \centering
    \small
        \renewcommand{\arraystretch}{0.9} 

    \caption{Facial recognition similarity score (lower the better) over different editing types. Three types of editing prompts are considered, including facial feature modifications, accessory adjustments, and background alterations.}
    \vspace{-0.8em}
    \begin{tabular}{@{}lccc@{}}
    \toprule
    \textbf{Method} & \textbf{Accessory} & \textbf{Facial Feature} & \textbf{Background} \\
    \midrule
    \textbf{No Defense} & 0.758 & 0.849 & 0.896 \\
    \midrule
    \textbf{PhotoGuard} & 0.507 & 0.483 & 0.837\\
    \textbf{EditShield} & 0.489 & 0.673 & 0.770\\
    \textbf{\ours} & \textbf{0.245} & \textbf{0.339} & \textbf{0.362}\\
    \bottomrule
    \end{tabular}
    \label{tab:edit-type}
    \vspace*{-1em}
\end{table}

\noindent \textbf{\ours~demonstrates superior robustness against common purification techniques compared to existing methods.}
We examined the robustness of the defense methods, we test three commonly used heuristic purification methods: Gaussian blurring, image rotation, and JPEG compression. These techniques were applied to images with adversarial perturbations. As shown in \textbf{Tab.~\ref{tab:robust}}, \ours~consistently outperforms both PhotoGuard and EditShield across all purification techniques.

\begin{table*}[t]
    \centering
    \small
        \renewcommand{\arraystretch}{0.9} 
    \caption{Robustness comparison of image protection methods against common purification techniques. Arrows $\uparrow$ and $\downarrow$ represent a higher or lower value is preferred for a successful defense. None denotes no purification techniques applied. Blur denotes Gaussian blurring $(k=5, \sigma=1.5)$, Rotate denotes random rotation between (-10, 10) degrees. JPEG $Q$ denotes JPEG compression at quality level $Q$.}
    \vspace{-0.8em}
    \resizebox{\textwidth}{!}{%
    \begin{tabular}{@{}lcccccccccccc@{}}
    \toprule
    & \multicolumn{6}{c}{\textbf{LPIPS} $\uparrow$} & \multicolumn{6}{c}{\textbf{FR} $\downarrow$} \\
    \cmidrule(lr){2-7} \cmidrule(l){8-13}
    \textbf{Method} & None & Blur & Rotate & JPEG 60 & JPEG 75 & JPEG 90 & None & Blur & Rotate & JPEG 60 & JPEG 75 & JPEG 90\\ 
    \midrule
    \textbf{PhotoGuard} & 0.376 & 0.306 & 0.367 & 0.249 & 0.248 & 0.278 & 0.523 & 0.804 & 0.719 & 0.786 & 0.782 & 0.780\\
    \textbf{EditShield} & 0.370 & 0.295 & 0.356 & 0.231 & 0.285 & 0.318 & 0.585 & 0.763 & 0.663 & 0.744 & 0.713 & 0.645\\
    \textbf{\ours} & \textbf{0.439} & \textbf{0.363} & \textbf{0.405} & \textbf{0.292} & \textbf{0.302} & \textbf{0.345} & \textbf{0.308} & \textbf{0.553} & \textbf{0.544} & \textbf{0.709} & \textbf{0.624} & \textbf{0.590}\\ 
    \bottomrule
    \end{tabular}%
    }
    \label{tab:robust}
    \vspace{-1.2em}
\end{table*}

\begin{table}[ht]
    \centering
    \small
    \renewcommand{\arraystretch}{0.9} 

    \caption{Facial recognition similarity score FR over different perturbation budgets. A lower FR is preferred for a successful defense.}
    \vspace{-0.8em}
    \begin{tabular}{@{}cccccc@{}}
        \toprule
        & \multicolumn{5}{c}{Perturbation Budgets}\\
        \cmidrule(lr){2-6}
        \textbf{Metrics} & 0.01 & 0.02 & 0.03 & 0.04 & 0.05 \\
        \midrule
        \textbf{FR} $\downarrow$ & 0.557 & 0.314 & 0.259 & 0.216 & 0.196\\
        \bottomrule
    \end{tabular}
    \label{tab:budget}
\end{table}

\begin{figure}[thb]
    \centering
    \begin{tabular}{cccc}
        \multicolumn{4}{c}{\footnotesize{\textbf{Editing Prompt}: `\textit{Let the person wear a police suit}'.}}\\
        \hspace*{-2mm}\includegraphics[width=0.23\linewidth]{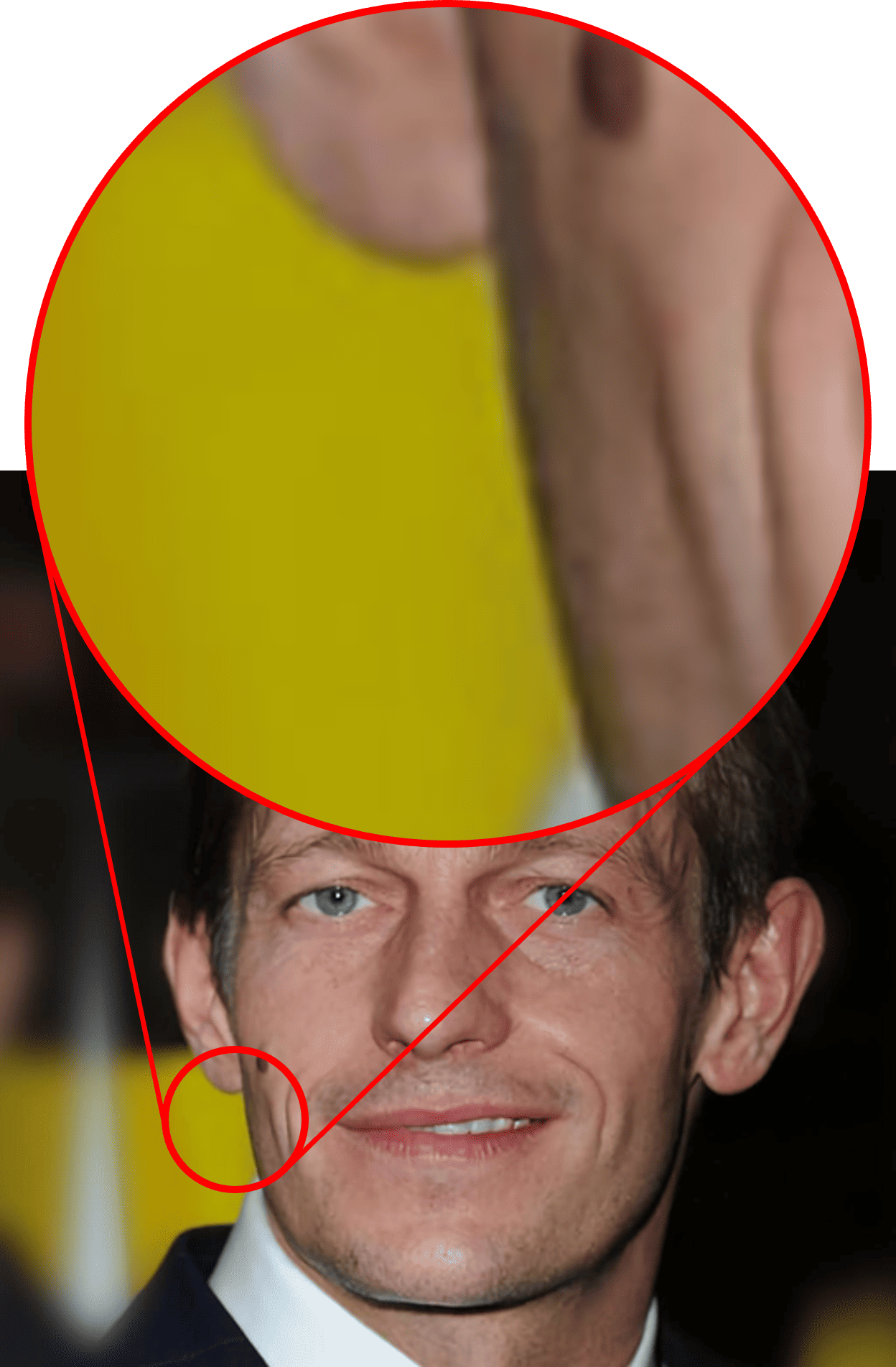}
        & \hspace*{-2mm}\includegraphics[width=0.23\linewidth]{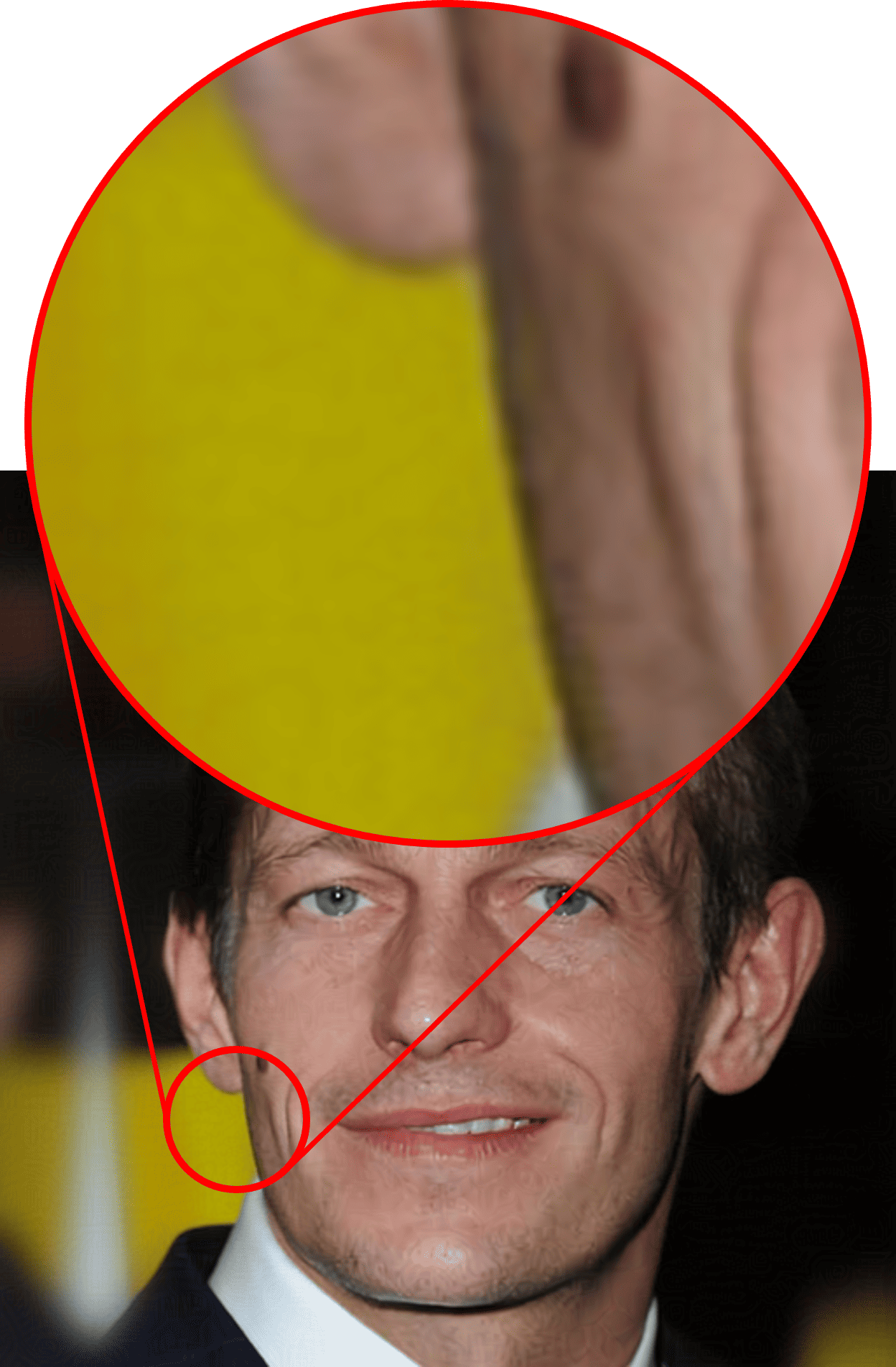}
        & \hspace*{-2mm}\includegraphics[width=0.23\linewidth]{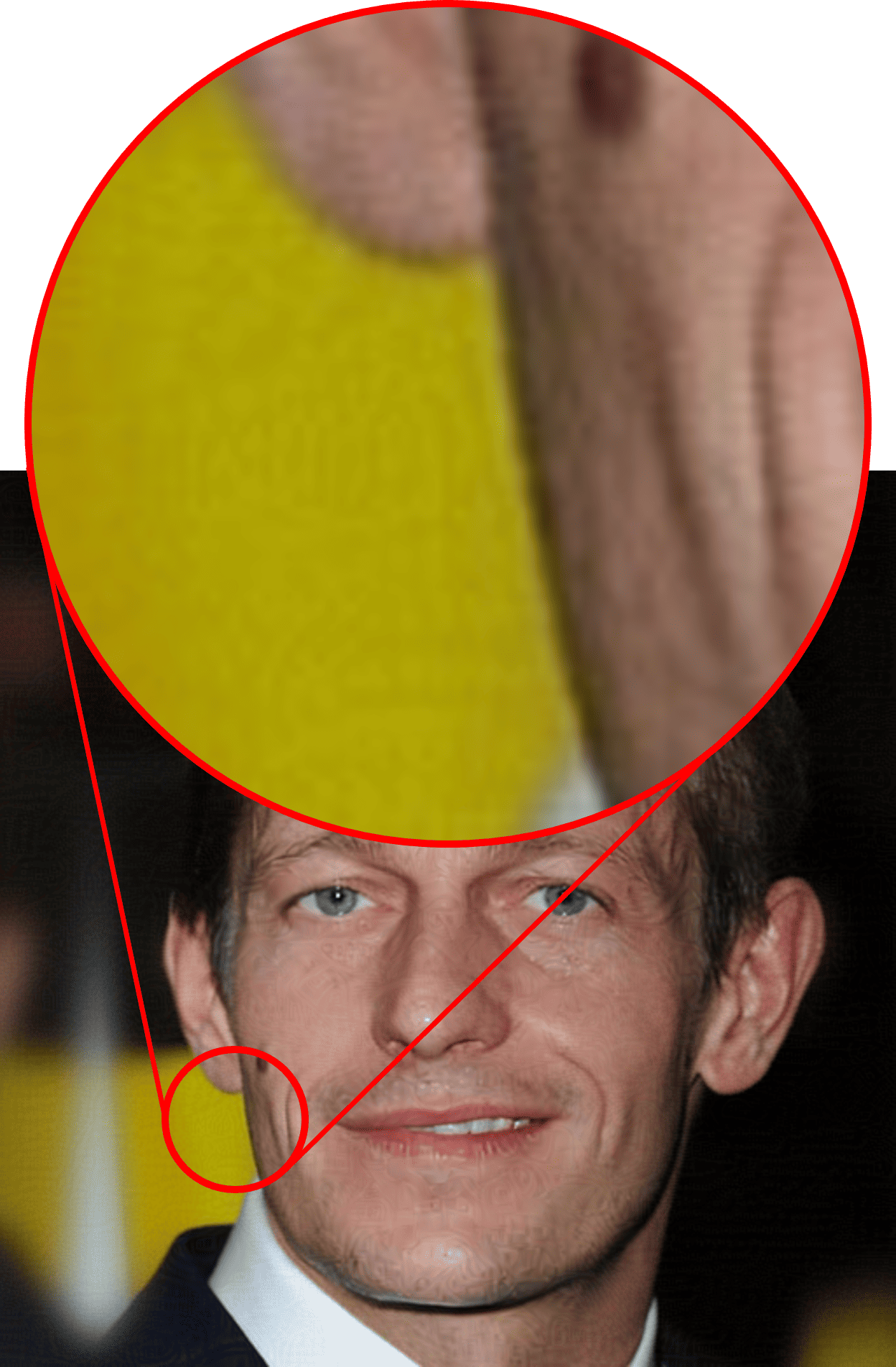}
        & \hspace*{-2mm}\includegraphics[width=0.23\linewidth]{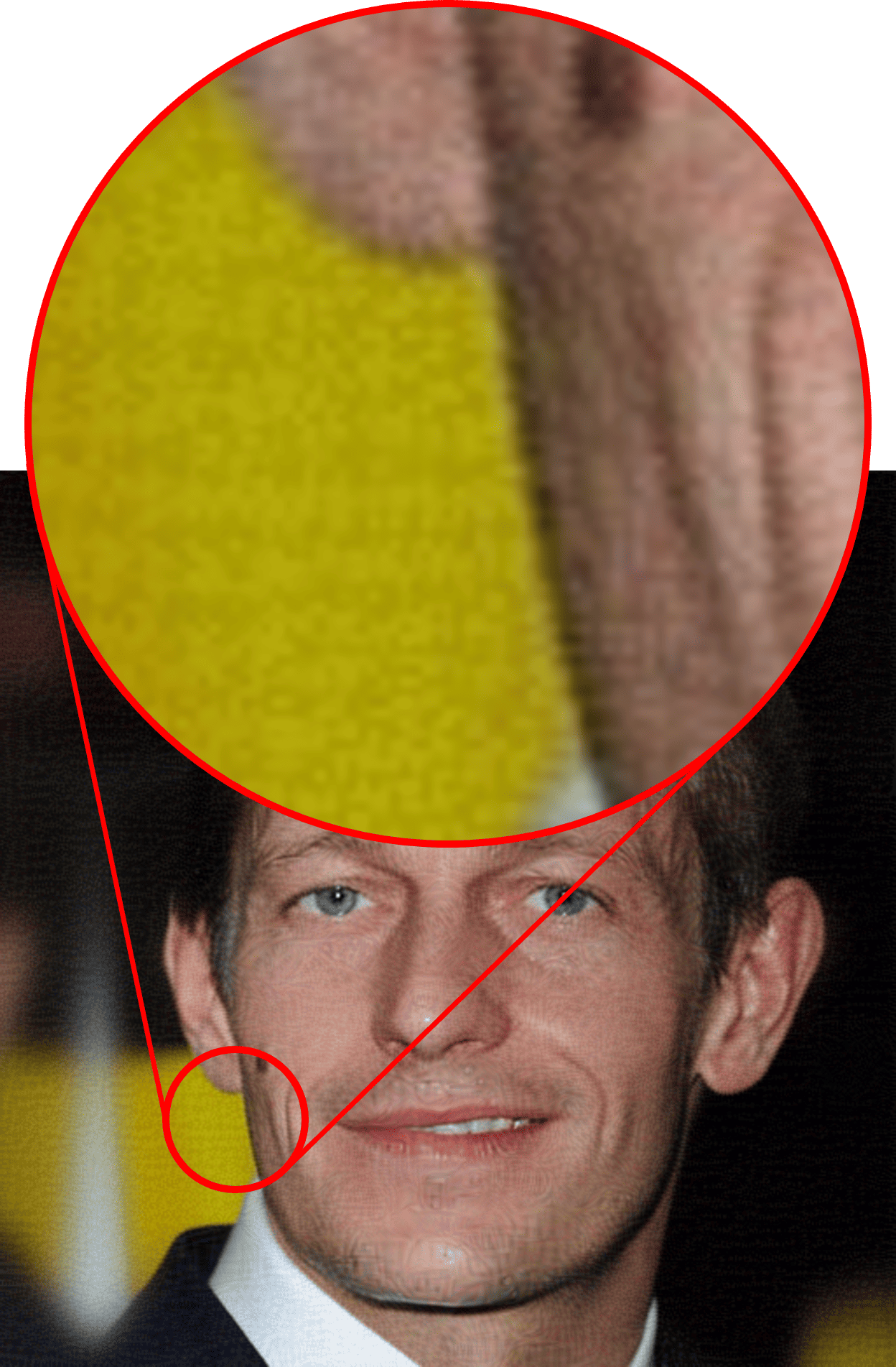}\\
        \hspace*{-2mm}\footnotesize{\textbf{(a) Budget=0}}
        & \hspace*{-2mm}\footnotesize{\textbf{(b) Budget=0.01}}
        & \hspace*{-2mm}\footnotesize{\textbf{(c) Budget=0.02}}
        & \hspace*{-2mm}\footnotesize{\textbf{(d) Budget=0.05}}\\
        \hspace*        {-2mm}\includegraphics[width=0.23\linewidth]{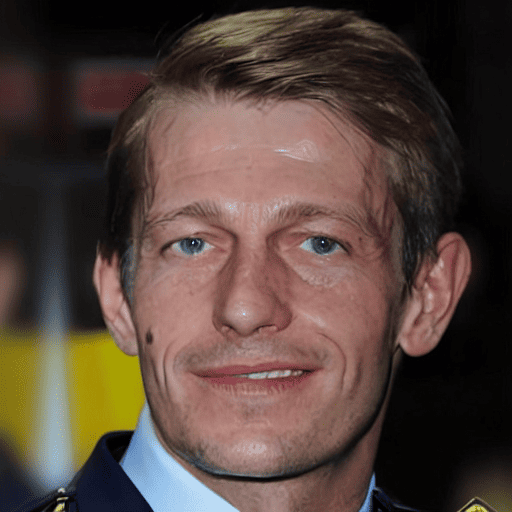}
        & \hspace*{-2mm}\includegraphics[width=0.23\linewidth]{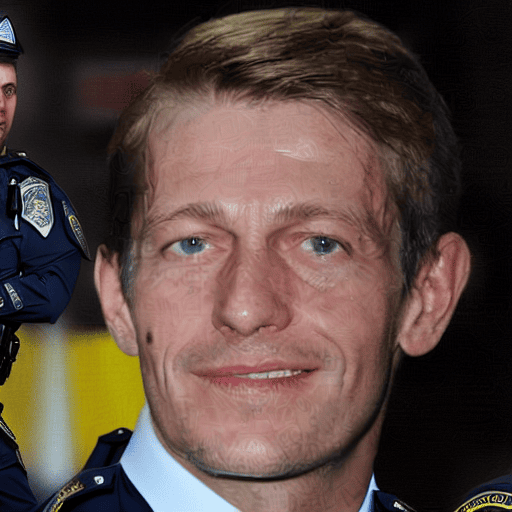}
        & \hspace*{-2mm}\includegraphics[width=0.23\linewidth]{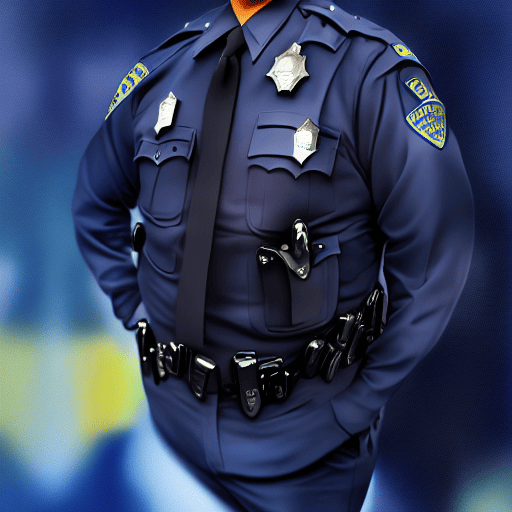}
        & \hspace*{-2mm}\includegraphics[width=0.23\linewidth]{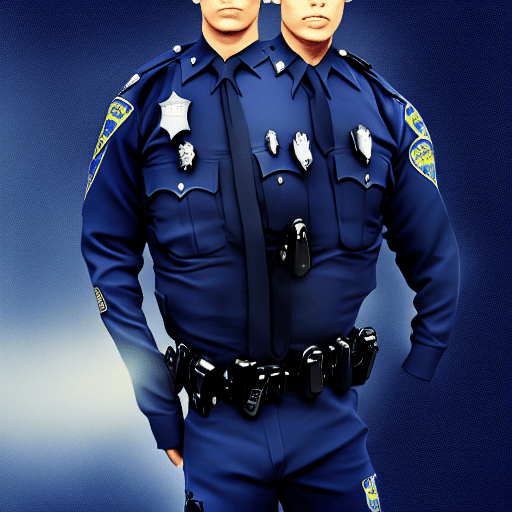}\\
    \end{tabular}
    \vspace*{-1em}
    \caption{Protected images injected with perturbation of different budgets, along with their corresponding editing results.}
    \label{fig:budget-example}
    \vspace*{-1em}
\end{figure}

\noindent\textbf{Ablation studies on perturbation budgets.} To demonstrate the impact of perturbation budgets on our protection method, we conducted an ablation study by varying the budget from 0.01 to 0.05. As shown in \textbf{Tab.~\ref{tab:budget}}, increasing the perturbation budget consistently reduces the facial recognition similarity score between the edit image and the source image, indicating stronger protection. However, as it is shown in \textbf{Fig.~\ref{fig:budget-example}}, large budgets (\eg, 0.05) introduce perceptible artifacts, compromising the image quality. Thus, we select a budget of 0.02 in our main experiments, which achieves effective protection with an imperceptible perturbation.

\begin{table}[ht]
    \centering
    \caption{Comparison of different design configurations and their impact on LPIPS and FR metrics. Arrows $\uparrow$ and $\downarrow$ represent a higher or lower value is preferred for a successful defense.}
    \vspace{-0.8em}
    \resizebox{\columnwidth}{!}{%
    \begin{tabular}{@{}lcccccc@{}}
    \toprule
    & \multicolumn{4}{c}{Components} & \multicolumn{2}{c}{Metrics}\\
    \cmidrule(lr){2-5} \cmidrule(lr){6-7}
    \textbf{Design} & CVL & Diffusion & Pixel & Feature & \textbf{LPIPS} $\uparrow$ & \textbf{FR} $\downarrow$\\
    \midrule
    \textbf{CVL} & \checkmark & & & & 0.091 & 0.667\\
    \textbf{CVL-D} & \checkmark & \checkmark & & & 0.381 & 0.380\\
    \textbf{CVL-DP} & \checkmark & \checkmark & \checkmark & & 0.381 & 0.573\\
    \textbf{\ours} & \checkmark & \checkmark & & \checkmark & \textbf{0.423} & \textbf{0.377}\\
    \bottomrule
    \end{tabular}%
    }
    \label{tab:design}
    \vspace*{-0.7em}
\end{table}

\noindent\textbf{Ablation studies on the effect of different protection components.} The analysis of protection components, as presented in \textbf{Tab.~\ref{tab:design}}, was conducted to evaluate the effectiveness of different elements in the perturbation optimization process. By examining various design configurations, we aimed to understand how each component contributes to the overall protection mechanism. The improvements in both the LPIPS metric and the FR metric from \textbf{Design I: CVL} to other design configurations showcases the importance of involving the diffusion process in the optimization loop. Interestingly, the LPIPS metric remains constant at 0.381 for both \textbf{Design II: CVL-D} and \textbf{Design III: CVL-DP} despite the addition of the pixel-level penalty in the optimization process. This observation aligns with our analysis in Sec.~\ref{sec: method}, underscoring that incorporating pixel-level loss does not disrupt the overall feature-level disparity. Furthermore, our proposed method \ours~achieves the best results in both LPIPS and FR metrics, indicating a better perceptual protection in both prompt fidelity and image integrity requirements.

\begin{wraptable}{r}{0.45\linewidth}
    \vspace*{-1em}
    \centering
    \small
        \renewcommand{\arraystretch}{0.9} 
    \caption{Performance comparison of {\ours} with different pretrained CNNs used as feature extractors.}
    \vspace{-0.8em}
    \resizebox{\linewidth}{!}{
    \begin{tabular}{lcc}
        \toprule
        \textbf{CNN} & \textbf{LPIPS} $\uparrow$ & \textbf{FR} $\downarrow$\\
        \midrule
        \textbf{AlexNet} & 0.451 & {0.346} \\
        \textbf{SqueezeNet} & 0.451 & {0.345} \\
        \textbf{VGG} & 0.458 & {0.340}\\
        \bottomrule
    \end{tabular}}
    \vspace*{-1em}
    \label{tab:cnn-robust}
\end{wraptable}

\noindent \textbf{\ours's robustness to feature extractor choices.} The results in \textbf{Tab.~\ref{tab:cnn-robust}} highlight that \ours~ performs consistently across different pretrained convolutional neural networks used as feature extractors. Specifically, we observe that the LPIPS scores, which measure the differences by comparing high-level semantic features of images are comparable between all three networks. Similarly, the FR scores, which assess the effectiveness of our method in disrupting biometric recognition, show that \ours~ achieves similar performance in reducing facial recognition similarity, regardless of the network used, reinforcing that \ours~remains robust in its protection across different feature extractors.

\noindent \textbf{Additional Results.} We also conducted additional experiments on evaluating the generalization abilities of \ours~protection on different datasets~\cite{karras2019style}. We refer more discussions in Appx.\,\ref{app: exp_results}.
\section{Conclusion, Limitation, and Discussion}
\label{sec:conclusion}
In this paper, we present \ours, an innovative method to protect human portrait images from malicious editing by optimizing adversarial perturbations that prevent biometric recognition post-editing. \ours~effectively disrupts identifiable facial features, breaking the biometric link between original and edited images. Experiments show its superior performance over existing defenses and robustness against purification techniques. While \ours~is tailored for single portraits, extending its efficacy to images with multiple individuals remains a challenge. Additionally, emerging generative models~\cite{tian2024visual,li20244k4dgen} like rectified flows~\cite{liu2022flow,rout2024semantic} may require further adaptations to sustain robustness. Addressing these challenges can enhance privacy protection at the forefront of generative models.

\section*{Broader Impact and Ethics Statement}
\noindent\textbf{Broader Impact Statement.} Advancements in diffusion-based image editing enable creative expression but also pose significant risks to privacy and identity security. Our work, {\ours}, addresses these risks by providing a robust defense mechanism that renders biometric information unrecognizable after edits. By demonstrating the potential pitfalls in current evaluation metrics, we aim to encourage the development of more reliable and effective solutions in this domain.

\noindent\textbf{Ethics Statement.} We believe our work sets a precedent for privacy-preserving AI research, especially in image synthesis and editing. By showing that privacy protection can be achieved through targeting biometric integrity, we hope to inspire more robust and innovative approaches to privacy in the broader context of Generative AI systems. This research also contributes to the ongoing dialogue about responsible AI development and highlights the importance of addressing privacy concerns as these technologies continue to advance.
 
\section*{Acknowledgment}
Y. Zhang and S. Liu were supported by the National Science Foundation (NSF) CISE Core Program Award IIS-2207052, the NSF Cyber-Physical Systems (CPS) Award CNS-2235231, the NSF CAREER Award IIS-2338068, the Cisco Research Award, and the Amazon Research Award for AI in Information Security. Y. Zhang was also partially supported by the IBM PhD Fellowship.

{
    \small
    \renewcommand{\bibname}{References}
    \bibliographystyle{IEEEtranN}
    \bibliography{main}

\begin{thebibliography}{66}
\providecommand{\natexlab}[1]{#1}
\providecommand{\url}[1]{#1}
\csname url@samestyle\endcsname
\providecommand{\newblock}{\relax}
\providecommand{\bibinfo}[2]{#2}
\providecommand{\BIBentrySTDinterwordspacing}{\spaceskip=0pt\relax}
\providecommand{\BIBentryALTinterwordstretchfactor}{4}
\providecommand{\BIBentryALTinterwordspacing}{\spaceskip=\fontdimen2\font plus
\BIBentryALTinterwordstretchfactor\fontdimen3\font minus \fontdimen4\font\relax}
\providecommand{\BIBforeignlanguage}[2]{{%
\expandafter\ifx\csname l@#1\endcsname\relax
\typeout{** WARNING: IEEEtranN.bst: No hyphenation pattern has been}%
\typeout{** loaded for the language `#1'. Using the pattern for}%
\typeout{** the default language instead.}%
\else
\language=\csname l@#1\endcsname
\fi
#2}}
\providecommand{\BIBdecl}{\relax}
\BIBdecl

\bibitem[Salman et~al.(2023)Salman, Khaddaj, Leclerc, Ilyas, and Madry]{salman2023raising}
H.~Salman, A.~Khaddaj, G.~Leclerc, A.~Ilyas, and A.~Madry, ``Raising the cost of malicious ai-powered image editing,'' \emph{arXiv preprint arXiv:2302.06588}, 2023.

\bibitem[Chen et~al.(2023)Chen, Jin, Liu, Chen, Wang, and Sun]{chen2023editshield}
R.~Chen, H.~Jin, Y.~Liu, J.~Chen, H.~Wang, and L.~Sun, ``Editshield: Protecting unauthorized image editing by instruction-guided diffusion models,'' \emph{arXiv preprint arXiv:2311.12066}, 2023.

\bibitem[Kawar et~al.(2023)Kawar, Zada, Lang, Tov, Chang, Dekel, Mosseri, and Irani]{kawar2023imagic}
B.~Kawar, S.~Zada, O.~Lang, O.~Tov, H.~Chang, T.~Dekel, I.~Mosseri, and M.~Irani, ``Imagic: Text-based real image editing with diffusion models,'' in \emph{Conference on Computer Vision and Pattern Recognition 2023}, 2023.

\bibitem[Zhang et~al.(2022{\natexlab{a}})Zhang, Han, Ghosh, Metaxas, and Ren]{zhang2022sine}
Z.~Zhang, L.~Han, A.~Ghosh, D.~Metaxas, and J.~Ren, ``Sine: Single image editing with text-to-image diffusion models,'' \emph{arXiv preprint arXiv:2212.04489}, 2022.

\bibitem[Zhang et~al.(2023{\natexlab{a}})Zhang, Mo, Chen, Sun, and Su]{Zhang2023MagicBrush}
K.~Zhang, L.~Mo, W.~Chen, H.~Sun, and Y.~Su, ``Magicbrush: A manually annotated dataset for instruction-guided image editing,'' in \emph{Advances in Neural Information Processing Systems}, 2023.

\bibitem[Hertz et~al.(2022)Hertz, Mokady, Tenenbaum, Aberman, Pritch, and Cohen-Or]{hertz2022prompt}
A.~Hertz, R.~Mokady, J.~Tenenbaum, K.~Aberman, Y.~Pritch, and D.~Cohen-Or, ``Prompt-to-prompt image editing with cross attention control,'' \emph{arXiv preprint arXiv:2208.01626}, 2022.

\bibitem[Huang et~al.(2024)Huang, Xie, Wang, Yuan, Cun, Ge, Zhou, Dong, Huang, Zhang, et~al.]{huang2024smartedit}
Y.~Huang, L.~Xie, X.~Wang, Z.~Yuan, X.~Cun, Y.~Ge, J.~Zhou, C.~Dong, R.~Huang, R.~Zhang \emph{et~al.}, ``Smartedit: Exploring complex instruction-based image editing with multimodal large language models,'' in \emph{Proceedings of the IEEE/CVF Conference on Computer Vision and Pattern Recognition}, 2024, pp. 8362--8371.

\bibitem[Fu et~al.(2024)Fu, Hu, Du, Wang, Yang, and Gan]{fu2024mgie}
T.-J. Fu, W.~Hu, X.~Du, W.~Y. Wang, Y.~Yang, and Z.~Gan, ``{Guiding Instruction-based Image Editing via Multimodal Large Language Models},'' in \emph{International Conference on Learning Representations (ICLR)}, 2024.

\bibitem[Brooks et~al.(2023)Brooks, Holynski, and Efros]{brooks2023instructpix2pix}
T.~Brooks, A.~Holynski, and A.~A. Efros, ``Instructpix2pix: Learning to follow image editing instructions,'' in \emph{Proceedings of the IEEE/CVF Conference on Computer Vision and Pattern Recognition}, 2023, pp. 18\,392--18\,402.

\bibitem[Choi et~al.(2023)Choi, Choi, Kim, Kim, and Yoon]{Choi2023CustomEditTI}
\BIBentryALTinterwordspacing
J.~Choi, Y.~Choi, Y.~Kim, J.~Kim, and S.-H. Yoon, ``Custom-edit: Text-guided image editing with customized diffusion models,'' \emph{ArXiv}, vol. abs/2305.15779, 2023. [Online]. Available: \url{https://api.semanticscholar.org/CorpusID:258888143}
\BIBentrySTDinterwordspacing

\bibitem[Li et~al.(2024{\natexlab{a}})Li, Li, Tu, Liu, Guo, Juefei-Xu, Xu, and Yu]{li2024light}
J.~Li, B.~Li, Z.~Tu, X.~Liu, Q.~Guo, F.~Juefei-Xu, R.~Xu, and H.~Yu, ``Light the night: A multi-condition diffusion framework for unpaired low-light enhancement in autonomous driving,'' in \emph{Proceedings of the IEEE/CVF Conference on Computer Vision and Pattern Recognition}, 2024, pp. 15\,205--15\,215.

\bibitem[Han et~al.(2023)Han, Yang, Kwon, and Ye]{han2023highlypersonalizedtextembedding}
\BIBentryALTinterwordspacing
I.~Han, S.~Yang, T.~Kwon, and J.~C. Ye, ``Highly personalized text embedding for image manipulation by stable diffusion,'' 2023. [Online]. Available: \url{https://arxiv.org/abs/2303.08767}
\BIBentrySTDinterwordspacing

\bibitem[Ruiz et~al.(2022)Ruiz, Li, Jampani, Pritch, Rubinstein, and Aberman]{ruiz2022dreambooth}
N.~Ruiz, Y.~Li, V.~Jampani, Y.~Pritch, M.~Rubinstein, and K.~Aberman, ``Dreambooth: Fine tuning text-to-image diffusion models for subject-driven generation,'' in \emph{Proceedings of the IEEE/CVF Conference on Computer Vision and Pattern Recognition (CVPR)}, 2022.

\bibitem[Gu et~al.(2023)Gu, Wang, Zhao, Fu, Xiong, Liu, Zhang, Zhang, Zhang, Jung, and Wang]{gu2023photoswap}
J.~Gu, Y.~Wang, N.~Zhao, T.-J. Fu, W.~Xiong, Q.~Liu, Z.~Zhang, H.~Zhang, J.~Zhang, H.~Jung, and X.~E. Wang, ``Photoswap: Personalized subject swapping in images,'' 2023.

\bibitem[Liu et~al.(2023)Liu, Huang, Chu, and Xu]{liu2023swapanything}
Z.~Liu, J.~Huang, H.~Chu, and Q.~Xu, ``Swapanything: Towards human-centric face and object swapping,'' \emph{IEEE Transactions on Pattern Analysis and Machine Intelligence}, 2023.

\bibitem[Qi et~al.(2024)Qi, Tu, Ye, Delbracio, Milanfar, Chen, and Talebi]{qi2024spire}
C.~Qi, Z.~Tu, K.~Ye, M.~Delbracio, P.~Milanfar, Q.~Chen, and H.~Talebi, ``Spire: Semantic prompt-driven image restoration,'' in \emph{European Conference on Computer Vision}.\hskip 1em plus 0.5em minus 0.4em\relax Springer, 2024, pp. 446--464.

\bibitem[Zohny et~al.(2023)Zohny, McMillan, and King]{zohny2023ethics}
H.~Zohny, J.~McMillan, and M.~King, ``Ethics of generative ai,'' pp. 79--80, 2023.

\bibitem[Vyas(2024)]{vyas2024ethical}
B.~Vyas, ``Ethical implications of generative ai in art and the media,'' \emph{International Journal for Multidisciplinary Research (IJFMR), E-ISSN}, pp. 2582--2160, 2024.

\bibitem[Lawton(2024)]{Lawton2024}
\BIBentryALTinterwordspacing
G.~Lawton. (2024) Generative ai ethics: 8 biggest concerns and risks. Published: 23 Jul 2024, Accessed: 2024-11-13. [Online]. Available: \url{https://www.techtarget.com/}
\BIBentrySTDinterwordspacing

\bibitem[Huang et~al.(2025)Huang, Gao, Wu, Wang, Wang, Zhou, Wang, Ye, Shi, Zhang, et~al.]{huang2025trustworthiness}
Y.~Huang, C.~Gao, S.~Wu, H.~Wang, X.~Wang, Y.~Zhou, Y.~Wang, J.~Ye, J.~Shi, Q.~Zhang \emph{et~al.}, ``On the trustworthiness of generative foundation models: Guideline, assessment, and perspective,'' \emph{arXiv preprint arXiv:2502.14296}, 2025.

\bibitem[Xing et~al.(2024)Xing, Hua, Gao, Zhu, Li, Tian, Li, Huang, Yang, Wang, et~al.]{xing2024autotrust}
S.~Xing, H.~Hua, X.~Gao, S.~Zhu, R.~Li, K.~Tian, X.~Li, H.~Huang, T.~Yang, Z.~Wang \emph{et~al.}, ``Autotrust: Benchmarking trustworthiness in large vision language models for autonomous driving,'' \emph{arXiv preprint arXiv:2412.15206}, 2024.

\bibitem[Times(2024)]{nyt_taylor_swift_2024}
\BIBentryALTinterwordspacing
T.~N.~Y. Times, ``Taylor swift ai fake images controversy,'' January 2024, accessed: 13-Nov-2024. [Online]. Available: \url{https://www.nytimes.com/2024/01/26/arts/music/taylor-swift-ai-fake-images.html}
\BIBentrySTDinterwordspacing

\bibitem[{BBC News}(2024)]{korean_deepfake_crisis_2024}
\BIBentryALTinterwordspacing
{BBC News}, ``Inside the deepfake porn crisis engulfing korean schools,'' September 2024, accessed: 13-Nov-2024. [Online]. Available: \url{https://www.bbc.com/news/articles/cpdlpj9zn9go}
\BIBentrySTDinterwordspacing

\bibitem[Zhao et~al.(2023)Zhao, Rao, Shi, Liu, Zhou, and Lu]{zhao2023diffswap}
W.~Zhao, Y.~Rao, W.~Shi, Z.~Liu, J.~Zhou, and J.~Lu, ``Diffswap: High-fidelity and controllable face swapping via 3d-aware masked diffusion,'' in \emph{Proceedings of the IEEE/CVF Conference on Computer Vision and Pattern Recognition}, 2023, pp. 8568--8577.

\bibitem[Tang et~al.(2019)Tang, Ma, Grobler, Meng, Wang, and Wen]{tang2019faces}
L.~Tang, W.~Ma, M.~Grobler, W.~Meng, Y.~Wang, and S.~Wen, ``Faces are protected as privacy: An automatic tagging framework against unpermitted photo sharing in social media,'' \emph{IEEE Access}, vol.~7, pp. 75\,556--75\,567, 2019.

\bibitem[An et~al.(2024)An, Zhang, Wu, Lin, Gu, and Wang]{an2024sd4privacy}
J.~An, W.~Zhang, D.~Wu, Z.~Lin, J.~Gu, and W.~Wang, ``Sd4privacy: exploiting stable diffusion for protecting facial privacy,'' in \emph{2024 IEEE International Conference on Multimedia and Expo (ICME)}.\hskip 1em plus 0.5em minus 0.4em\relax IEEE, 2024, pp. 1--6.

\bibitem[He et~al.(2024)He, Zhu, Chen, Wang, and Gao]{he2024diff}
X.~He, M.~Zhu, D.~Chen, N.~Wang, and X.~Gao, ``Diff-privacy: Diffusion-based face privacy protection,'' \emph{IEEE Transactions on Circuits and Systems for Video Technology}, 2024.

\bibitem[Shan et~al.(2023)Shan, Wenger, Zhang, Li, Zheng, and Zhao]{shan2023glaze}
S.~Shan, E.~Wenger, J.~Zhang, H.~Li, H.~Zheng, and B.~Y. Zhao, ``Glaze: Protecting artists from style mimicry by text-to-image models,'' \emph{USENIX Security Symposium}, 2023.

\bibitem[Wang et~al.(2023)Wang, Chang, Gandikota, and Jha]{wang2023distraction}
R.~Wang, H.~Chang, D.~Gandikota, and S.~Jha, ``Distraction is all you need: Instruction-based image editing with complementary attention,'' \emph{arXiv preprint arXiv:2306.05934}, 2023.

\bibitem[Huang and Zhao(2023)]{huang2023nightshade}
E.~Huang and B.~Y. Zhao, ``Nightshade: A data poisoning tool to protect artists from generative ai,'' \emph{arXiv preprint arXiv:2310.13828}, 2023.

\bibitem[Liang et~al.(2023)Liang, Wu, Hua, Zhang, Xue, Song, Xue, Ma, and Guan]{liang2023adversarial}
C.~Liang, X.~Wu, Y.~Hua, J.~Zhang, Y.~Xue, T.~Song, Z.~Xue, R.~Ma, and H.~Guan, ``Adversarial example does good: preventing painting imitation from diffusion models via adversarial examples,'' in \emph{Proceedings of the 40th International Conference on Machine Learning}, 2023, pp. 20\,763--20\,786.

\bibitem[Zhang et~al.(2023{\natexlab{b}})Zhang, Xu, Cui, Meng, Wu, and Lyu]{zhang2023robustness}
J.~Zhang, Z.~Xu, S.~Cui, C.~Meng, W.~Wu, and M.~R. Lyu, ``On the robustness of latent diffusion models,'' \emph{arXiv preprint arXiv:2306.08257}, 2023.

\bibitem[Rombach et~al.(2022)Rombach, Blattmann, Lorenz, Esser, and Ommer]{rombach2022high}
R.~Rombach, A.~Blattmann, D.~Lorenz, P.~Esser, and B.~Ommer, ``High-resolution image synthesis with latent diffusion models,'' in \emph{Proceedings of the IEEE/CVF conference on computer vision and pattern recognition}, 2022, pp. 10\,684--10\,695.

\bibitem[Nesterov(2013)]{nesterov2013introductory}
Y.~Nesterov, \emph{Introductory lectures on convex optimization: A basic course}.\hskip 1em plus 0.5em minus 0.4em\relax Springer Science \& Business Media, 2013, vol.~87.

\bibitem[Carlini and Wagner(2017)]{carlini2017towards}
N.~Carlini and D.~Wagner, ``Towards evaluating the robustness of neural networks,'' in \emph{IEEE Symposium on Security and Privacy (SP)}, 2017.

\bibitem[Kim et~al.(2022)Kim, Jain, and Yu]{kim2022adaface}
M.~Kim, A.~K. Jain, and X.~Yu, ``Adaface: Quality adaptive margin for face recognition,'' in \emph{Proceedings of the IEEE/CVF Conference on Computer Vision and Pattern Recognition}, 2022.

\bibitem[Deng et~al.(2019)Deng, Guo, Xue, and Zafeiriou]{deng2019arcface}
J.~Deng, J.~Guo, N.~Xue, and S.~Zafeiriou, ``Arcface: Additive angular margin loss for deep face recognition,'' in \emph{Proceedings of the IEEE/CVF Conference on Computer Vision and Pattern Recognition}, 2019.

\bibitem[Wang et~al.(2018)Wang, Wang, Zhou, Ji, Gong, Zhou, Li, and Liu]{wang2018cosface}
H.~Wang, Y.~Wang, Z.~Zhou, X.~Ji, D.~Gong, J.~Zhou, Z.~Li, and W.~Liu, ``Cosface: Large margin cosine loss for deep face recognition,'' in \emph{Proceedings of the IEEE Conference on Computer Vision and Pattern Recognition}, 2018.

\bibitem[Boutros et~al.(2022{\natexlab{a}})Boutros, Damer, Kirchbuchner, and Kuijper]{Boutros_2022_CVPR}
F.~Boutros, N.~Damer, F.~Kirchbuchner, and A.~Kuijper, ``Elasticface: Elastic margin loss for deep face recognition,'' in \emph{Proceedings of the IEEE/CVF Conference on Computer Vision and Pattern Recognition (CVPR) Workshops}, June 2022, pp. 1578--1587.

\bibitem[Huang et~al.(2020)Huang, Wang, Tai, Liu, Shen, Li, and Jilin~Li]{huang2020curricularface}
Y.~Huang, Y.~Wang, Y.~Tai, X.~Liu, P.~Shen, S.~Li, and F.~H. Jilin~Li, ``Curricularface: Adaptive curriculum learning loss for deep face recognition,'' in \emph{Proceedings of the IEEE/CVF Conference on Computer Vision and Pattern Recognition (CVPR) Workshops}, 2020, pp. 1--8.

\bibitem[Terh{\"{o}}rst et~al.(2021)Terh{\"{o}}rst, Ihlefeld, Huber, Damer, Kirchbuchner, Raja, and Kuijper]{QMagFace}
\BIBentryALTinterwordspacing
P.~Terh{\"{o}}rst, M.~Ihlefeld, M.~Huber, N.~Damer, F.~Kirchbuchner, K.~Raja, and A.~Kuijper, ``{QMagFace}: Simple and accurate quality-aware face recognition,'' \emph{CoRR}, vol. abs/2111.13475, 2021. [Online]. Available: \url{https://arxiv.org/abs/2111.13475}
\BIBentrySTDinterwordspacing

\bibitem[Meng et~al.(2021)Meng, Zhao, Huang, and Zhou]{meng2021magface}
Q.~Meng, S.~Zhao, Z.~Huang, and F.~Zhou, ``{MagFace}: A universal representation for face recognition and quality assessment,'' in \emph{CVPR}, 2021.

\bibitem[Boutros et~al.(2022{\natexlab{b}})Boutros, Huber, Siebke, Rieber, and Damer]{Sface_Boutros}
\BIBentryALTinterwordspacing
F.~Boutros, M.~Huber, P.~Siebke, T.~Rieber, and N.~Damer, ``Sface: Privacy-friendly and accurate face recognition using synthetic data,'' in \emph{{IEEE} International Joint Conference on Biometrics, {IJCB} 2022, Abu Dhabi, United Arab Emirates, October 10-13, 2022}.\hskip 1em plus 0.5em minus 0.4em\relax {IEEE}, 2022, pp. 1--11. [Online]. Available: \url{https://doi.org/10.1109/IJCB54206.2022.10007961}
\BIBentrySTDinterwordspacing

\bibitem[Boutros et~al.(2024)Boutros, Huber, Luu, Siebke, and Damer]{10454585}
F.~Boutros, M.~Huber, A.~T. Luu, P.~Siebke, and N.~Damer, ``Sface2: Synthetic-based face recognition with w-space identity-driven sampling,'' \emph{IEEE Transactions on Biometrics, Behavior, and Identity Science}, pp. 1--1, 2024.

\bibitem[Boutros et~al.(2023)Boutros, Klemt, Fang, Kuijper, and Damer]{BoutrosKFKD23}
\BIBentryALTinterwordspacing
F.~Boutros, M.~Klemt, M.~Fang, A.~Kuijper, and N.~Damer, ``Unsupervised face recognition using unlabeled synthetic data,'' in \emph{17th {IEEE} International Conference on Automatic Face and Gesture Recognition, {FG} 2023, Waikoloa Beach, HI, USA, January 5-8, 2023}.\hskip 1em plus 0.5em minus 0.4em\relax {IEEE}, 2023, pp. 1--8. [Online]. Available: \url{https://doi.org/10.1109/FG57933.2023.10042627}
\BIBentrySTDinterwordspacing

\bibitem[Kolf et~al.(2023)Kolf, Rieber, Elliesen, Boutros, Kuijper, and Damer]{10208805}
\BIBentryALTinterwordspacing
J.~N. Kolf, T.~Rieber, J.~Elliesen, F.~Boutros, A.~Kuijper, and N.~Damer, ``{ Identity-driven Three-Player Generative Adversarial Network for Synthetic-based Face Recognition },'' in \emph{2023 IEEE/CVF Conference on Computer Vision and Pattern Recognition Workshops (CVPRW)}.\hskip 1em plus 0.5em minus 0.4em\relax Los Alamitos, CA, USA: IEEE Computer Society, Jun. 2023, pp. 806--816. [Online]. Available: \url{https://doi.ieeecomputersociety.org/10.1109/CVPRW59228.2023.00088}
\BIBentrySTDinterwordspacing

\bibitem[Qiu et~al.(2021)Qiu, Yu, Gong, Li, Liu, and Tao]{qiu2021synface}
H.~Qiu, B.~Yu, D.~Gong, Z.~Li, W.~Liu, and D.~Tao, ``Synface: Face recognition with synthetic data,'' in \emph{Proceedings of the IEEE/CVF International Conference on Computer Vision}, 2021, pp. 10\,880--10\,890.

\bibitem[Nie et~al.(2022)Nie, Guo, Huang, Xiao, Vahdat, and Anandkumar]{nie2022diffusion}
W.~Nie, B.~Guo, Y.~Huang, C.~Xiao, A.~Vahdat, and A.~Anandkumar, ``Diffusion models for adversarial purification,'' \emph{arXiv preprint arXiv:2205.07460}, 2022.

\bibitem[Wang et~al.(2004)Wang, Bovik, Sheikh, and Simoncelli]{wang2004image}
Z.~Wang, A.~C. Bovik, H.~R. Sheikh, and E.~P. Simoncelli, ``Image quality assessment: from error visibility to structural similarity,'' \emph{IEEE transactions on image processing}, vol.~13, no.~4, pp. 600--612, 2004.

\bibitem[Radford et~al.(2021)Radford, Kim, Hallacy, Ramesh, Goh, Agarwal, Sastry, Askell, Mishkin, Clark, et~al.]{radford2021learning}
A.~Radford, J.~W. Kim, C.~Hallacy, A.~Ramesh, G.~Goh, S.~Agarwal, G.~Sastry, A.~Askell, P.~Mishkin, J.~Clark \emph{et~al.}, ``Learning transferable visual models from natural language supervision,'' in \emph{International conference on machine learning}.\hskip 1em plus 0.5em minus 0.4em\relax PMLR, 2021, pp. 8748--8763.

\bibitem[Zhang et~al.(2024{\natexlab{a}})Zhang, Zhang, Yao, Jia, Liu, Liu, and Liu]{zhang2024unlearncanvas}
Y.~Zhang, Y.~Zhang, Y.~Yao, J.~Jia, J.~Liu, X.~Liu, and S.~Liu, ``Unlearncanvas: A stylized image dataset to benchmark machine unlearning for diffusion models,'' \emph{arXiv preprint arXiv:2402.11846}, 2024.

\bibitem[Tu et~al.(2021)Tu, Wang, Birkbeck, Adsumilli, and Bovik]{tu2021ugc}
Z.~Tu, Y.~Wang, N.~Birkbeck, B.~Adsumilli, and A.~C. Bovik, ``Ugc-vqa: Benchmarking blind video quality assessment for user generated content,'' \emph{IEEE Transactions on Image Processing}, vol.~30, pp. 4449--4464, 2021.

\bibitem[Zhang et~al.(2018)Zhang, Isola, Efros, Shechtman, and Wang]{zhang2018unreasonable}
R.~Zhang, P.~Isola, A.~A. Efros, E.~Shechtman, and O.~Wang, ``The unreasonable effectiveness of deep features as a perceptual metric,'' in \emph{Proceedings of the IEEE conference on computer vision and pattern recognition}, 2018, pp. 586--595.

\bibitem[Karras(2017)]{karras2017progressive}
T.~Karras, ``Progressive growing of gans for improved quality, stability, and variation,'' \emph{arXiv preprint arXiv:1710.10196}, 2017.

\bibitem[Zhang et~al.(2022{\natexlab{b}})Zhang, Zhang, Khanduri, Hong, Chang, and Liu]{zhang2022revisiting}
Y.~Zhang, G.~Zhang, P.~Khanduri, M.~Hong, S.~Chang, and S.~Liu, ``Revisiting and advancing fast adversarial training through the lens of bi-level optimization,'' in \emph{International Conference on Machine Learning}.\hskip 1em plus 0.5em minus 0.4em\relax PMLR, 2022, pp. 26\,693--26\,712.

\bibitem[Zhang et~al.(2022{\natexlab{c}})Zhang, Zhang, Zhang, Fan, Li, Liu, and Chang]{zhang2022fairness}
G.~Zhang, Y.~Zhang, Y.~Zhang, W.~Fan, Q.~Li, S.~Liu, and S.~Chang, ``Fairness reprogramming,'' \emph{Advances in Neural Information Processing Systems}, vol.~35, pp. 34\,347--34\,362, 2022.

\bibitem[Zhang et~al.(2022{\natexlab{d}})Zhang, Lu, Zhang, Chen, Chen, Fan, Martie, Horesh, Hong, and Liu]{zhang2022distributed}
G.~Zhang, S.~Lu, Y.~Zhang, X.~Chen, P.-Y. Chen, Q.~Fan, L.~Martie, L.~Horesh, M.~Hong, and S.~Liu, ``Distributed adversarial training to robustify deep neural networks at scale,'' in \emph{Uncertainty in artificial intelligence}.\hskip 1em plus 0.5em minus 0.4em\relax PMLR, 2022, pp. 2353--2363.

\bibitem[Zhang et~al.(2023{\natexlab{c}})Zhang, Cai, Chen, Zhang, Zhang, Chen, Chang, Wang, and Liu]{zhang2023robust}
Y.~Zhang, R.~Cai, T.~Chen, G.~Zhang, H.~Zhang, P.-Y. Chen, S.~Chang, Z.~Wang, and S.~Liu, ``Robust mixture-of-expert training for convolutional neural networks,'' in \emph{Proceedings of the IEEE/CVF International Conference on Computer Vision}, 2023, pp. 90--101.

\bibitem[Zhang et~al.(2023{\natexlab{d}})Zhang, Jia, Chen, Chen, Zhang, Liu, Ding, and Liu]{zhang2023generate}
Y.~Zhang, J.~Jia, X.~Chen, A.~Chen, Y.~Zhang, J.~Liu, K.~Ding, and S.~Liu, ``To generate or not? safety-driven unlearned diffusion models are still easy to generate unsafe images... for now,'' \emph{arXiv preprint arXiv:2310.11868}, 2023.

\bibitem[Zhang et~al.(2024{\natexlab{b}})Zhang, Chen, Jia, Zhang, Fan, Liu, Hong, Ding, and Liu]{zhang2024defensive}
Y.~Zhang, X.~Chen, J.~Jia, Y.~Zhang, C.~Fan, J.~Liu, M.~Hong, K.~Ding, and S.~Liu, ``Defensive unlearning with adversarial training for robust concept erasure in diffusion models,'' \emph{arXiv preprint arXiv:2405.15234}, 2024.

\bibitem[Zhuang et~al.(2023)Zhuang, Zhang, and Liu]{zhuang2023pilot}
H.~Zhuang, Y.~Zhang, and S.~Liu, ``A pilot study of query-free adversarial attack against stable diffusion,'' in \emph{Proceedings of the IEEE/CVF Conference on Computer Vision and Pattern Recognition}, 2023, pp. 2385--2392.

\bibitem[Karras et~al.(2019)Karras, Laine, and Aila]{karras2019style}
T.~Karras, S.~Laine, and T.~Aila, ``A style-based generator architecture for generative adversarial networks,'' in \emph{Proceedings of the IEEE/CVF conference on computer vision and pattern recognition}, 2019, pp. 4401--4410.

\bibitem[Tian et~al.(2024)Tian, Jiang, Yuan, Peng, and Wang]{tian2024visual}
K.~Tian, Y.~Jiang, Z.~Yuan, B.~Peng, and L.~Wang, ``Visual autoregressive modeling: Scalable image generation via next-scale prediction,'' \emph{Advances in neural information processing systems}, vol.~37, pp. 84\,839--84\,865, 2024.

\bibitem[Li et~al.(2024{\natexlab{b}})Li, Pan, Yang, Xu, Zhou, Zhang, Li, Kadambi, Wang, Tu, et~al.]{li20244k4dgen}
R.~Li, P.~Pan, B.~Yang, D.~Xu, S.~Zhou, X.~Zhang, Z.~Li, A.~Kadambi, Z.~Wang, Z.~Tu \emph{et~al.}, ``4k4dgen: Panoramic 4d generation at 4k resolution,'' \emph{arXiv preprint arXiv:2406.13527}, 2024.

\bibitem[Liu et~al.(2022)Liu, Gong, and Liu]{liu2022flow}
X.~Liu, C.~Gong, and Q.~Liu, ``Flow straight and fast: Learning to generate and transfer data with rectified flow,'' \emph{arXiv preprint arXiv:2209.03003}, 2022.

\bibitem[Rout et~al.(2024)Rout, Chen, Ruiz, Caramanis, Shakkottai, and Chu]{rout2024semantic}
L.~Rout, Y.~Chen, N.~Ruiz, C.~Caramanis, S.~Shakkottai, and W.-S. Chu, ``Semantic image inversion and editing using rectified stochastic differential equations,'' \emph{arXiv preprint arXiv:2410.10792}, 2024.

\end{thebibliography}
}
\clearpage
\newpage
\onecolumn

\section*{\Large{Appendix}}
\setcounter{section}{0}
\setcounter{figure}{0}
\setcounter{table}{0}
\makeatletter 
\renewcommand{\thesection}{\Alph{section}}
\renewcommand{\theHsection}{\Alph{section}}
\renewcommand{\thefigure}{A\arabic{figure}} 
\renewcommand{\theHfigure}{A\arabic{figure}} 
\renewcommand{\thetable}{A\arabic{table}}
\renewcommand{\theHtable}{A\arabic{table}}
\makeatother

\renewcommand{\thetable}{A\arabic{table}}
\setcounter{equation}{0}
\renewcommand{\theequation}{A\arabic{equation}}

\appendix

\section{Detailed Experiment Setups}
\label{app: exp_setup}
\subsection{Implementation Details of {\ours}}
{\ours} optimizes perturbation on facial disruption and feature embedding disparity that prevent biometric recognition post-editing. The pseudocode of {\ours} is presented in Algorithm~\ref{alg:ours}. More specifically, the facial recognition loss function $f_{FR}$ is defined as the negative of the similarity score between the input images computed by the {\cvl} model\footnote{The model is available on \href{https://github.com/mk-minchul/CVLface}{https://github.com/mk-minchul/CVLface}}, and the feature disparity loss function $f_{FE}$ is computed as the weighted sum of the layer-wise feature embedding distances across the feature extractor network. As mentioned in Sec \ref{sec: experiments}, we also include the untargeted latent-wise loss from EditShield\cite{chen2023editshield} as a regularization term to stabilize the protection results. The hyper-parameters used in our implementation are summarized in \textbf{Tab.~\ref{tab:attack-hyper}}.

\begin{algorithm}[ht]
\caption{{\ours}}
\label{alg:ours}
\begin{algorithmic}[1]
\Require Input image ${\bx}$, VAE $\mathcal{E}, \mathcal{D}$ in the diffusion model, step size $\alpha$, number of steps $N$, overall perturbation budget $\epsilon$, regularization weight $\lambda$, facial recognition loss function $f_\text{FR}$, feature disparity loss function $f_\text{FE}$
\State Initialize perturbation $\bdelta\gets N(0, \textbf{I})$, and the protected image $\bx'\gets\bx+\bdelta$
\State Compute the latent embedding of the input image $\bz\gets\mathcal{E}(\bx)$
\For{$n=1$ to $N$}
    \State Compute the latent embedding of the protected image $\bz'\gets\mathcal{E}(\bx')$
    \State Compute the decoded image from the latent embedding $\bx_d\gets\mathcal{D}(\bz')$
    \State Compute the facial recognition loss $l_\text{FR}\gets f_\text{FR}(\bx_d, \bx)$
    \State Compute the feature disparity loss $l_\text{FE}\gets f_\text{FE}(\bx_d, \bx)$
    \State Compute the latent loss (regularization term) $l_\text{L}\gets\lVert\bz'-\bz\rVert_2^2$
    \State Update the perturbation $\bdelta\gets\bdelta + \alpha\cdot\text{sign}(\nabla_{\bx'}(l_\text{FR}+l_\text{FE}+\lambda\cdot l_\text{L}))$
    \State $\bdelta\gets\text{clip}(\bdelta, -\epsilon, \epsilon)$
    \State Update the protected image: $\bx'\gets\bx+\bdelta$
\EndFor
\Ensure The protected image $\bx'$
\end{algorithmic}
\end{algorithm}

\begin{table}[ht]
\centering
\small
\caption{Hyper-parameters used for the implementation.}
\label{tab:attack-hyper}
\vspace{-0.8em}
\begin{tabular}{ccccc}
    \toprule
    \textbf{Norm} & \textbf{perturbation budget} $\epsilon$ & \textbf{step size} $\alpha$ & \textbf{number of steps} $N$ & $\lambda$ \\
    \midrule
    $l_\infty$ & 0.02 & 0.003 & 100 & 0.2\\
    \bottomrule
\end{tabular}
\end{table}

\subsection{Implementation Details of Baselines}
In addition to using previous methods~\cite{salman2023raising, chen2023editshield} as baselines, we also compare our {\ours} approach against several widely used techniques in the adversarial machine learning field. These methods are summarized in Algorithms~\ref{alg:un}, \ref{alg:vae}, and~\ref{alg:cw}. To ensure a fair comparison, we use the same hyper-parameters settings in Tab. \ref{tab:attack-hyper}.

\begin{algorithm}[ht]
\caption{Untargeted Encoder Attack}
\label{alg:un}
\begin{algorithmic}[1]
\Require Input image ${\bx}$, VAE $\mathcal{E}$ in the diffusion model, step size $\alpha$, number of steps $N$, overall perturbation budget $\epsilon$
\State Initialize perturbation $\bdelta\gets N(0, \textbf{I})$, and the protected image $\bx'\gets\bx+\bdelta$
\State Compute the latent embedding of the input image $\bz\gets\mathcal{E}(\bx)$
\For{$n=1$ to $N$}
    \State Compute the latent embedding of the protected image $\bz'\gets\mathcal{E}(\bx')$
    \State Compute the latent loss $l\gets\lVert\bz'-\bz\rVert_2^2$
    \State Update the perturbation $\bdelta\gets\bdelta + \alpha\cdot\text{sign}(\nabla_{\bx'}l)$
    \State $\bdelta\gets\text{clip}(\bdelta, -\epsilon, \epsilon)$
    \State Update the protected image $\bx'\gets\bx+\bdelta$
\EndFor
\Ensure The protected image $\bx'$
\end{algorithmic}
\end{algorithm}

\begin{algorithm}[ht]
\caption{VAE Attack}
\label{alg:vae}
\begin{algorithmic}[1]
\Require Input image ${\bx}$, target image $\bx_\text{tgt}$,VAE $\mathcal{E}$, $\mathcal{D}$ in the diffusion model, step size $\alpha$, number of steps $N$, overall perturbation budget $\epsilon$
\State Initialize perturbation $\bdelta\gets N(0, \textbf{I})$, and the protected image $\bx'\gets\bx+\bdelta$
\For{$n=1$ to $N$}
    \State Compute the decoded image $\bx_d\gets\mathcal{D}(\mathcal{E}(\bx'))$ 
    \State Compute the loss $l\gets\lVert\bx_d-\bx_\text{tgt}\rVert_2^2$
    \State Update the perturbation $\bdelta\gets\bdelta - \alpha\cdot\text{sign}(\nabla_{\bx'}l)$
    \State $\bdelta\gets\text{clip}(\bdelta, -\epsilon, \epsilon)$
    \State Update the protected image $\bx'\gets\bx+\bdelta$
\EndFor
\Ensure The protected image $\bx'$
\end{algorithmic}
\end{algorithm}

\begin{algorithm}[ht]
\caption{CW $L_2$ Attack}
\label{alg:cw}
\begin{algorithmic}[1]
\Require Input image ${\bx}$, VAE $\mathcal{E}$ in the diffusion model, step size $\alpha$, number of steps $N$, overall perturbation budget $\epsilon$, weight $c$
\State Initialize $\bw\gets\mathbf{0}$
\State Compute the latent embedding of the input image: $\bz\gets\mathcal{E}(\bx)$
\For{$n = 1$ to $N$}
    \State Compute the protected image $\bx'\gets\frac{1}{2}(\tanh(\bw)+1)$
    \State Compute the latent embedding of the protected image $\bz'\gets\mathcal{E}(\bx')$
    \State Compute the $L_2$ loss $l_{L_2}\gets\lVert\bx'-\bx\rVert_2^2$
    \State Compute the latent loss $l_\text{L}\gets-\lVert\bz'-\bz\rVert_2^2$
    \State Update $\bw\gets\bw-\alpha\cdot\nabla_{\bw}(l_{L_2}+c\cdot l_{L})$
\EndFor
\State Compute $\bdelta\gets\text{clip}(\frac{1}{2}(\tanh(\bw)+1) - \bx, -\epsilon, \epsilon)$
\State Compute the protected image $\bx'\gets\bx+\bdelta$
\Ensure The protected image $\bx'$
\end{algorithmic}
\end{algorithm}

\subsection{Image Editing Details}
\noindent\textbf{Models.} For image editing, we use the open-source instruction-guided diffusion model InstructPix2Pix~\cite{brooks2023instructpix2pix} hosted on Hugging Face\footnote{The model is available on \href{https://huggingface.co/timbrooks/instruct-pix2pix}{https://huggingface.co/timbrooks/instruct-pix2pix}} as our primary target model. We use the hyper-parameters presented in \textbf{Tab.~\ref{tab:edit-hyper}}. We use the same seed setting when comparing edits on the unprotected images and the images protected by different methods to ensure that the edit images are are modified in the same way and that the different editing effects are due to the protection methods instead of random seeds.

\noindent\textbf{Dataset.} For the human portrait images used in our experiments, we utilize a filtered subset of the CelebA-HQ dataset\footnote{The dataset is available on \href{https://www.kaggle.com/datasets/lamsimon/celebahq/data}{https://www.kaggle.com/datasets/lamsimon/celebahq/data}}, a high-quality human face attribute dataset widely used in the facial analysis community. The dataset consists of $2,000$ human portrait images ensuring diversity across various demographic groups, including race, age, and gender, to enhance the representativeness of our experiments. For the editing prompts, we manually selected $25$ prompts across three categories: facial feature modification, accessory adjustments, and background alternations. These prompts were specifically selected to produce noticeable changes across a wide range of images, avoiding those that would fail to affect a certain subset (\eg, \textquotedblleft\textit{Let the person wear glasses}\textquotedblright~ will be ineffective for individuals who already wear glasses, which is a significant portion of the dataset). The specific prompts utilized in our experiments are listed in \textbf{Tab.~\ref{tab:editing_prompts}} for detailed reference.

\begin{table}[ht]
\centering
\small
\caption{Hyper-parameters used for the image editing process.}
\label{tab:edit-hyper}
\vspace{-0.8em}
\begin{tabular}{cccc}
    \toprule
    \textbf{image size} & \textbf{inference steps} & \textbf{image guidance scale} & \textbf{text guidance scale} \\
    \midrule
    512$\times$512 & 50 & 1.5 & 7.5\\
    \bottomrule
\end{tabular}
\end{table}

\begin{table}[ht]
\centering
\small
\caption{Editing prompts categorized into facial feature modifications, accessory adjustments, and background alterations.}
\label{tab:editing_prompts}
\vspace{-0.8em}
\begin{tabular}{p{4cm}p{12cm}}
\toprule
\textbf{Category} & \textbf{Prompts} \\ 
\midrule
\multirow{4}{*}{Facial Feature Modifications} & \ding{182} Turn the person's hair pink; \ding{183} Let the person turn bald; \ding{184} Let the person have a tattoo; \ding{185} Let the person wear purple makeup; \ding{186} Let the person grow a mustache; \ding{187} Turn the person into a zombie; \ding{188} Change the skin color to Avatar blue; \ding{189} Add elf-like ears; \ding{190} Add large vampire fangs; \ding{191} Apply Goth style makeup. \\ 
\midrule
\multirow{3}{*}{Accessory Adjustments} &  \ding{182} Let the person wear a police suit; \ding{183} Let the person wear a bowtie; \ding{184} Let the person wear a helmet; \ding{185} Let the person wear sunglasses; \ding{186} Let the person wear earrings; \ding{187} Let the person smoke a cigar; \ding{188} Place a headband in the hair; \ding{189} Place a tiara on the top of the head.\\ 
\midrule
\multirow{3}{*}{Background Alterations} & \ding{182} Let it be snowy; \ding{183} Change the background to a beach; \ding{184} Add a city skyline background; \ding{185} Add a forest background; \ding{186} Change the background to a desert; \ding{187} Set the background in a library; \ding{188} Let the person stand under the moon;\\
\bottomrule
\end{tabular}
\end{table}

\subsection{Evaluation Metrics}
\noindent\textbf{PSNR, SSIM, and LPIPS scores.} In our experiments, we compute the PSNR and SSIM scores using the torchmetrics library\footnote{This library can be installed from \href{https://lightning.ai/docs/torchmetrics/stable/}{https://lightning.ai/docs/torchmetrics/stable/}}, while the LPIPS score is computed using the lpips library\footnote{This library can be installed from \href{https://pypi.org/project/lpips/}{https://pypi.org/project/lpips/}}. All these three metrics are computed by comparing the similarity between the edited image without defense and the edited image with defense. A lower similarity score (lower PSNR, SSIM score and higher LPIPS score) indicates better protection. PSNR and SSIM primarily focus on pixel-level statistical information, while LPIPS evaluates the similarity of high-level semantic features, capturing perceptual differences that are more aligned with human visual perception.

\noindent\textbf{CLIP-S score.} In the main paper, we utilize the CLIP-S metric to assess the prompt fidelity by computing the similarity between the image embedding shift and the text embedding in the CLIP embedding space:
{
\small
\vspace{-0.5em}
\begin{equation}
    \text{CLIP-S}=\frac{(E_\text{edit}-E_\text{src})\cdot E_\text{prompt}}{\lVert E_\text{edit}-E_\text{src}\rVert\lVert E_\text{prompt}\rVert},
\end{equation}
}

\noindent where $E_\text{src}$ denotes the CLIP image embedding of the source image, $E_{edit}$ denotes the CLIP image embedding of the edited image, and $E_\text{prompt}$ denotes the CLIP text embedding of the prompt instruction. This formulation is particularly suitable for our experiments because the prompts are designed as instructions describing the expected transformation or modification from the source image to the edited image.

\noindent\textbf{CLIP-SD score.} Following PhotoGuard's evaluation metric~\cite{salman2023raising}, an alternative approach to assess the prompt fidelity is to compute the cosine similarity directly between the embedding of the edited image and the embedding of the descriptive text prompt in the CLIP embedding space:
{
\small
\vspace{-0.5em}
\begin{equation}
    \text{CLIP-SD}=\frac{E_\text{edit}\cdot E_\text{desc}}{\lVert E_\text{edit}\rVert\lVert E_\text{desc}\rVert},
\end{equation}
}

\noindent where $E_\text{desc}$ denotes the CLIP text embedding of the descriptive text prompt. We report the CLIP-SD score for each method in \textbf{Tab.~\ref{tab:CLIP-SD}}. From the table, we observe that, except for the VAE method, all defense methods show a worse defense effect compared to the \textquotedblleft{No Defense}\textquotedblright~scenario. This aligns with the analysis presented in Sec~\ref{sec:evaluation}, where we discussed how CLIP-based similarity metrics often overemphasize the elements from the prompt, leading to a prioritization of over-editing. To generate the descriptive text prompts, we leverage ChatGPT based on the prompt instructions provided in Tab.~\ref{tab:editing_prompts}.

\begin{table}[ht]
    \centering
    \small
    \caption{Quantitative evaluation on prompt fidelity using CLIP-SD. The $\downarrow$ indicates that a lower CLIP-SD score is preferred for a successful defense.}
    \label{tab:CLIP-SD}
    \vspace{-0.8em}
    \resizebox{\textwidth}{!}{%
    \begin{tabular}{cccccccc}
        \toprule
        \textbf{Method} & \textbf{No Defense} & \textbf{PhotoGuard} & \textbf{EditShield} & \textbf{Untargeted Encoder} & \textbf{CW L2} & \textbf{VAE} & \textbf{\ours(ours)} \\
        \midrule
        \textbf{CLIP-SD}$\downarrow$ & 0.272\footnotesize{$\pm$0.029} & 0.283\footnotesize{$\pm0.029$} & 0.277\footnotesize{$\pm$0.027} & 0.284\footnotesize{$\pm$0.024} & 0.277\footnotesize{$\pm$0.027} & 0.270\footnotesize{$\pm$0.029} & 0.283\footnotesize{$\pm$0.024} \\
        \bottomrule
    \end{tabular}%
    }
\end{table}

\noindent\textbf{CLIP-I score.} In the main paper, we utilize the CLIP-I metric to assess the image integrity by computing the similarity between the edited image embedding and the source image embedding in the CLIP embedding space:
{
\small
\vspace{-0.5em}
\begin{equation}
    \text{CLIP-I}=\frac{E_\text{edit}\cdot E_\text{src}}{\lVert E_\text{edit}\rVert\lVert E_\text{src}\rVert}.
\end{equation}
}
\noindent The CLIP-I metric is used as a general indicator of the preservation effect, providing an overall measure of how similar the edited image is to the source image in the CLIP embedding space. While this serves as a useful first step in generally evaluating image integrity, it does not specifically address biometric integrity, which is central to protecting human portrait images.

\noindent\textbf{FR score.} In the main paper, we utilize the {\cvl} model to compute the facial recognition similarity score between the edited and source image to indicate the preservation effect of biometric integrity:
{
\small
\vspace{-0.5em}
\begin{equation}
    \text{FR}=\text{\cvl}(I_\text{edit}, I_\text{src}),
\end{equation}
}

\noindent where $I_\text{src}$ denotes the source image, and $I_\text{edit}$ denotes the edited image. Unlike other general image similarity metrics, the {\cvl} model is tailored to assess the consistency of facial features, making it more suitable for evaluating how well the identity of the person is preserved after the image has been edited. The FR score plays a key role in assessing whether the protection method effectively disrupts the biometric identity of the person in the image.

\newpage
\section{Additional Experiment Results}
\label{app: exp_results}
\subsection{Qualitative Results on Background Alternation}
\begin{figure}[h]
    \centering
    \resizebox{0.8\textwidth}{!}{%
        \includegraphics{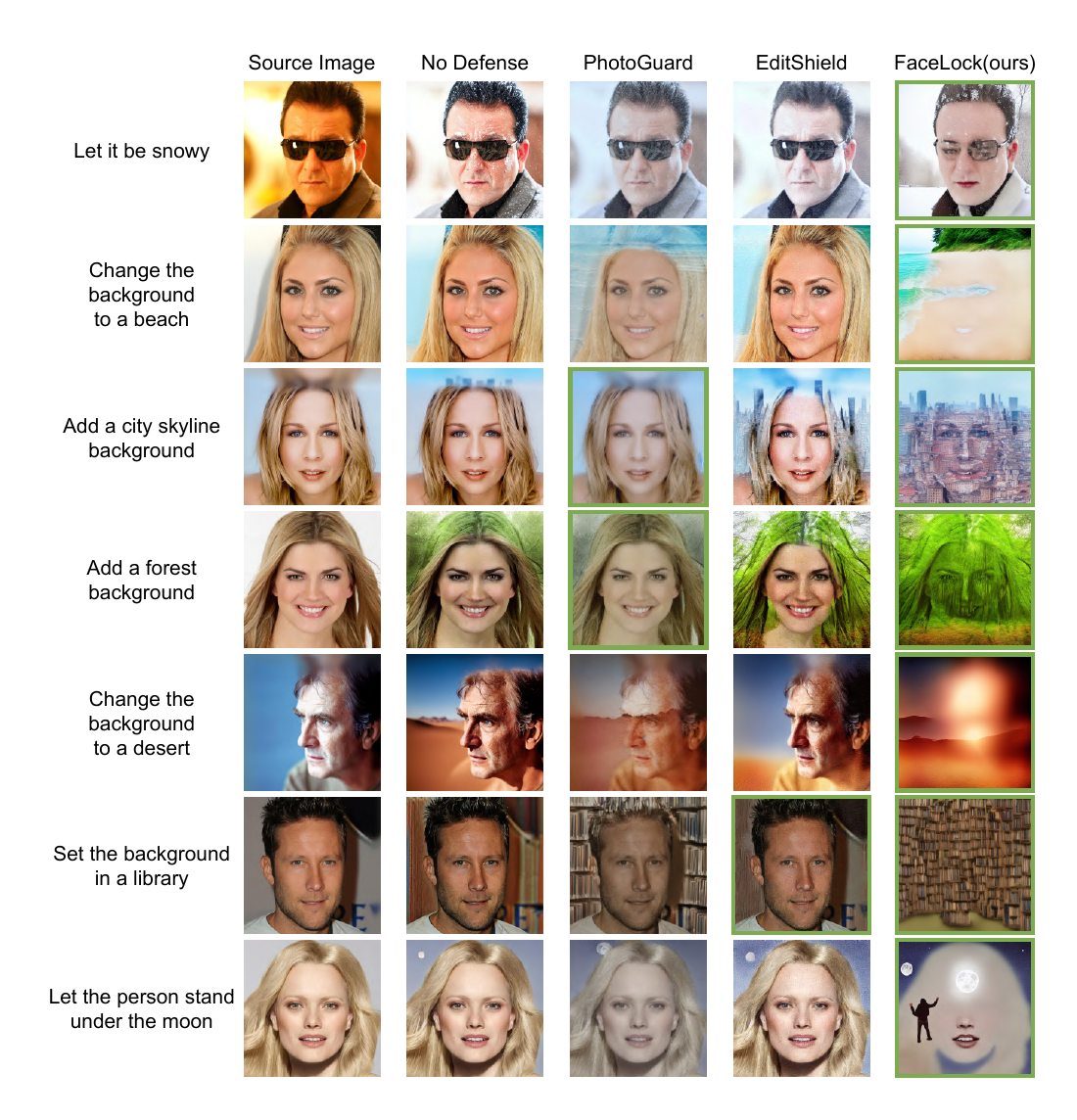}
    }
    \vspace{-0.8em}
    \caption{Qualitative results of background alternation edits across various defense methods. Images in green frames denote successful defense.}
    \label{fig:background}
\end{figure}

\newpage
\subsection{Qualitative Results on Accessory Adjustment}
\begin{figure}[h]
    \centering
    \resizebox{0.8\textwidth}{!}{%
        \includegraphics{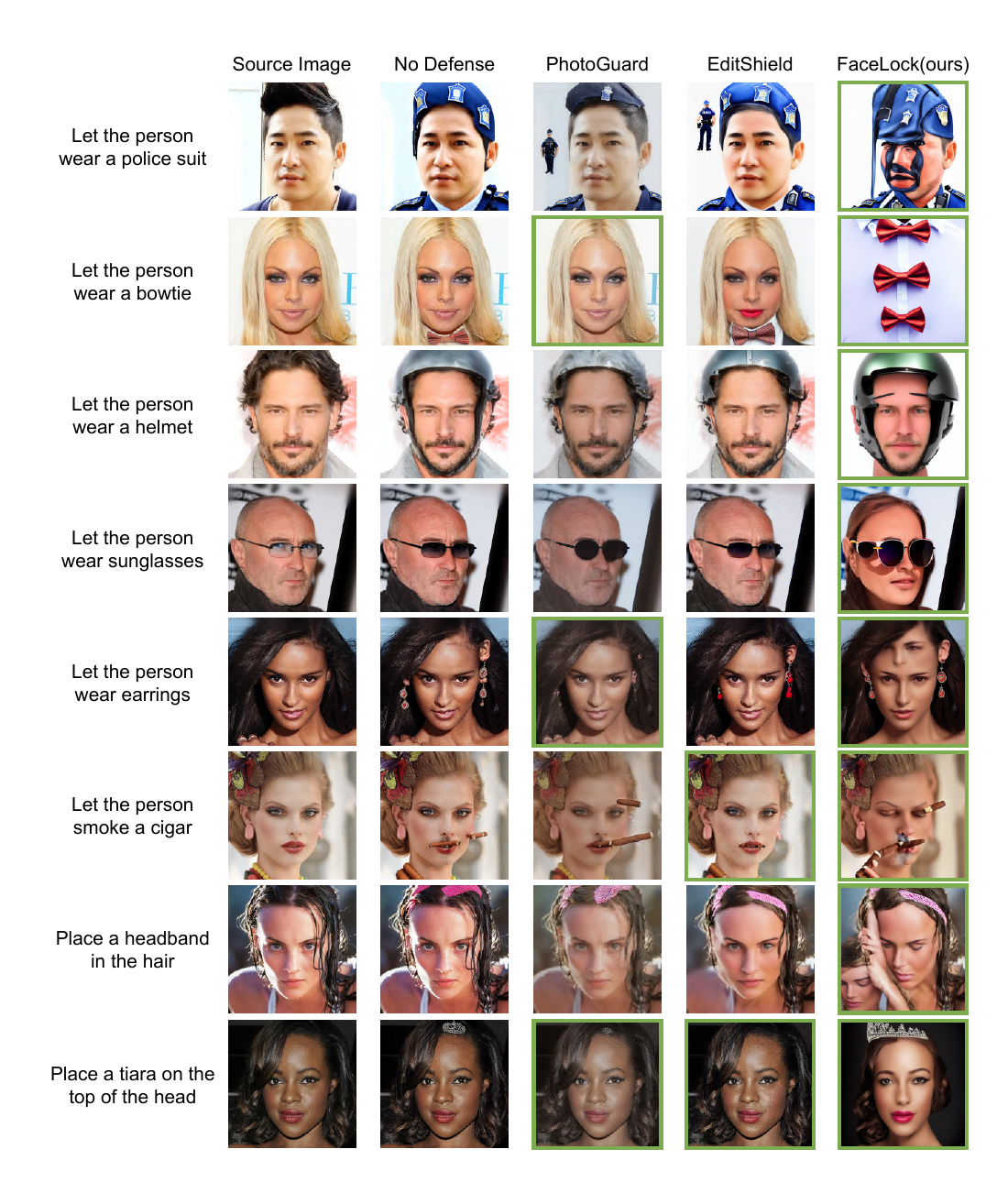}
    }
    \vspace{-0.8em}
    \caption{Qualitative results of accessory adjustment edits across various defense methods. Images in green frames denote successful defense.}
    \label{fig:accessory}
\end{figure}

\newpage
\subsection{Qualitative Results on Facial Feature Modification}
\begin{figure}[h]
    \centering
    \resizebox{0.8\textwidth}{!}{%
        \includegraphics{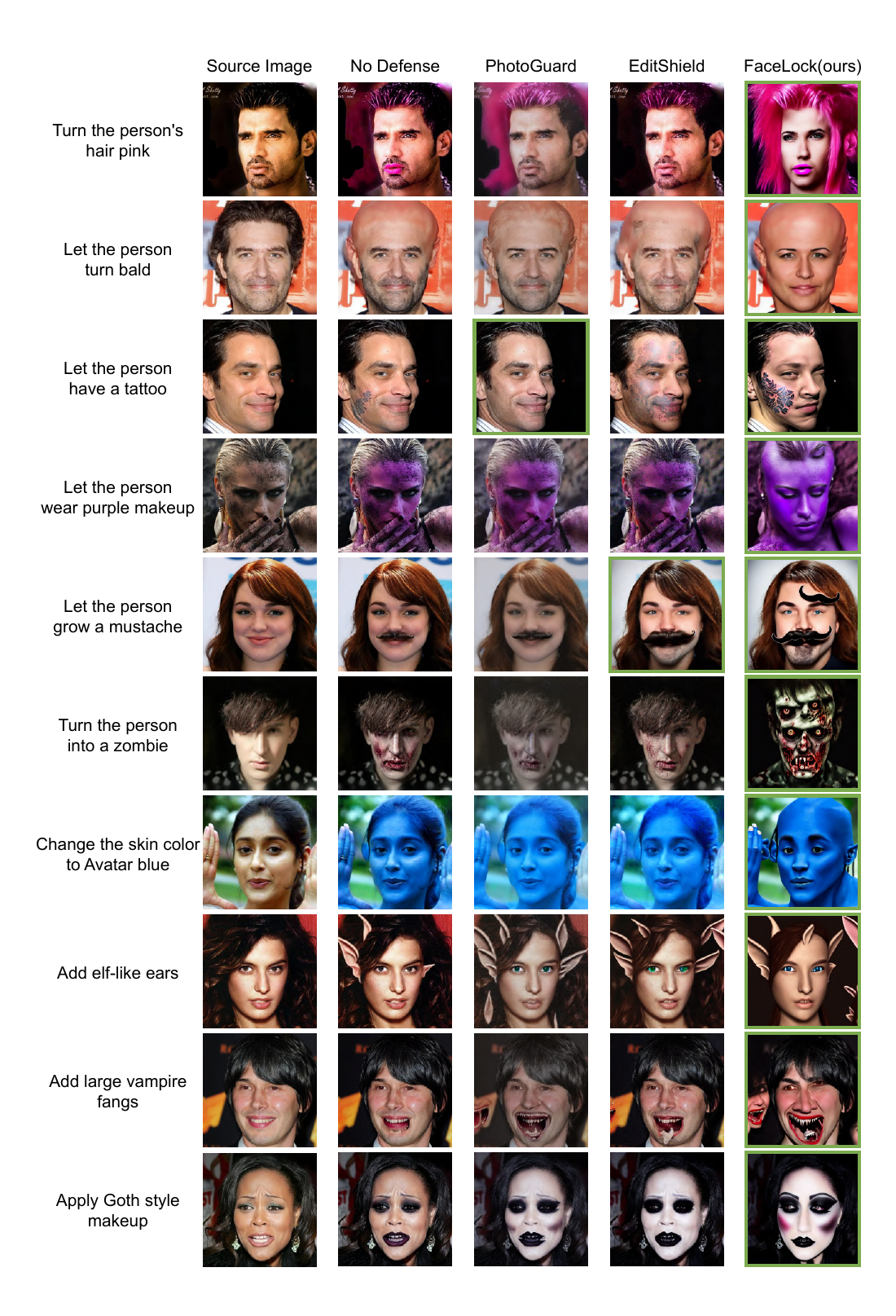}
    }
    \vspace{-0.8em}
    \caption{Qualitative results of facial feature modification edits across various defense methods. Images in green frames denote successful defense.}
    \label{fig:facial}
\end{figure}

\newpage
\subsection{Qualitative Results Against Purification}
\begin{figure}[h]
    \centering
    \resizebox{0.78\textwidth}{!}{%
        \includegraphics{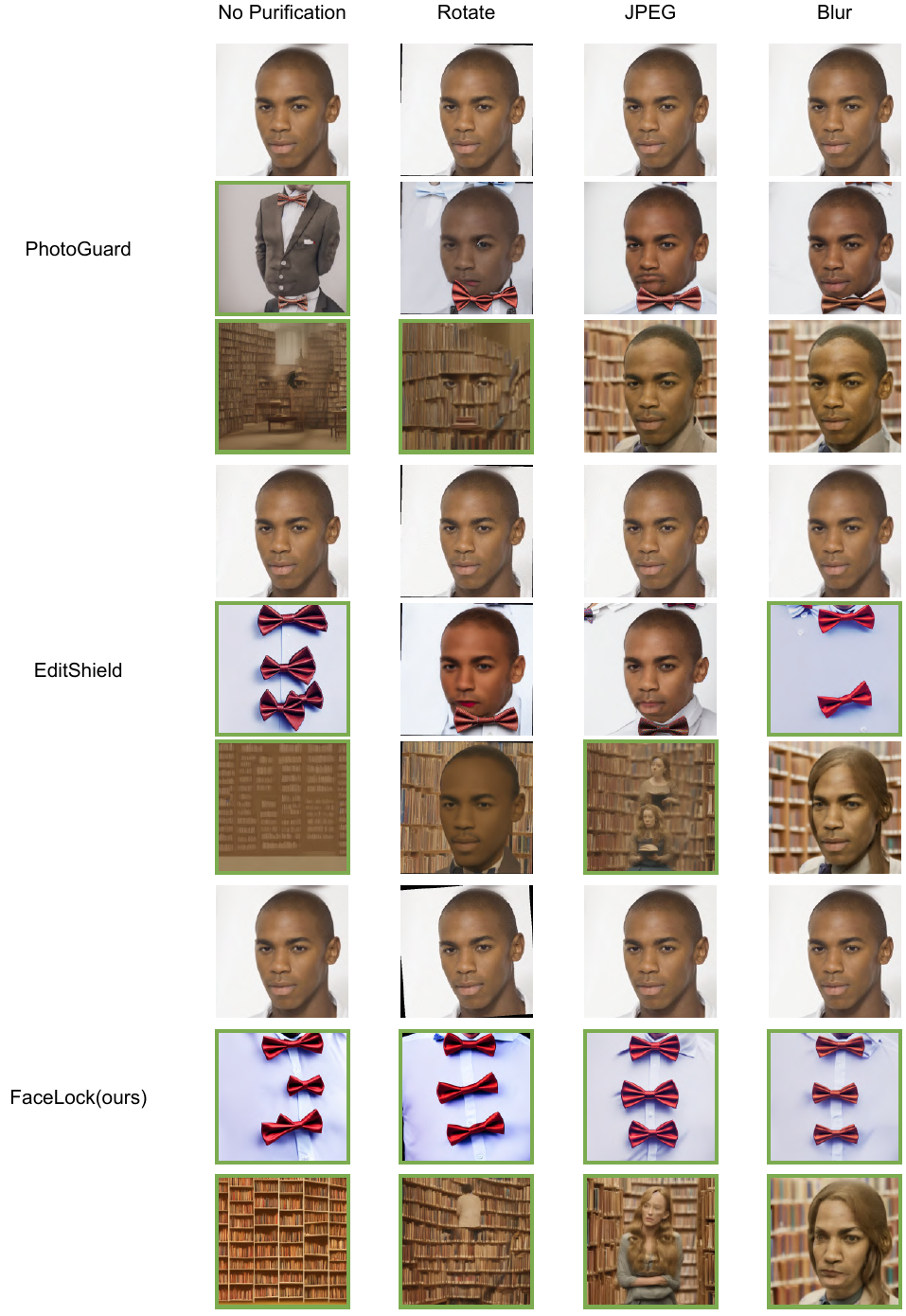}
    }
    \vspace{-0.8em}
    \caption{Qualitative results of edits on protected images after applying purification methods. Each block shows: purified protected images (\nth{1} row), edits with the instruction \textquotedblleft Let the person wear a bowtie\textquotedblright, and edits with the instruction \textquotedblleft Set the background in a library\textquotedblright. Purification methods include random rotation (-10, 10), JPEG compression (quality 75), and Gaussian blurring ($k=5, \sigma=1.5$). Images in green frames denote successful defense.}
    \label{fig:robust}
\end{figure}

\subsection{Results on Other Datasets}
To further evaluate the effectiveness of \ours, we compare its performance against existing baselines on a subset of the Flickr-Faces-HQ (FFHQ) dataset\cite{karras2019style}. As shown in \textbf{Tab.~\ref{tab:ffhq}}, \ours achieves the lowest FR score of 0.356, demonstrating its strong identity protection while maintaining competitive performance across other key metrics.

\begin{table}[h]
    \centering
    \small
    \caption{Quantitative evaluation on the FFHQ dataset.}
    \vspace{-0.8em}
    \begin{tabular}{lcccccc}
        \toprule
        \textbf{Method} & $\textbf{CLIP-S}\downarrow$ & $\textbf{PSNR}\downarrow$ & $\textbf{SSIM}\downarrow$ & $\textbf{LPIPS}\uparrow$ & $\textbf{CLIP-I}\downarrow$ & $\textbf{FR}\downarrow$ \\
        \midrule
        \textbf{No Defense} & 0.108 & - & - & - & 0.860 & 0.820 \\
        \midrule
        \textbf{PhotoGuard} & \textbf{0.095} & \textbf{14.76} & 0.555 & 0.515 & 0.681 & 0.449 \\
        \textbf{EditShield} & 0.098 & 18.92 & \textbf{0.532} & 0.480 & 0.753 & 0.633 \\
        \textbf{\ours} & 0.099 & 17.02 & 0.538 & \textbf{0.542} & \textbf{0.680} & \textbf{0.356}\\
        \bottomrule
    \end{tabular}
    \label{tab:ffhq}
\end{table}

\subsection{Results on Other Purification Methods}
\begin{table}[h]
    \centering
    \small
    \caption{Robustness comparison against other purification methods.}
    \vspace{-0.8em}
    \begin{tabular}{lcccc}
        \toprule
        & \multicolumn{2}{c}{\textbf{LPIPS} $\uparrow$} & \multicolumn{2}{c}{\textbf{FR} $\downarrow$} \\
        \cmidrule(lr){2-3} \cmidrule(lr){4-5} 
        \textbf{Method} & Color Jitter & DiffPure & Color Jitter & DiffPure\\
        \midrule
        \textbf{PhotoGuard} & 0.275 & 0.311 & 0.686 & 0.691\\
        \textbf{EditShield} & 0.303 & \textbf{0.316} & 0.593 & 0.610\\
        \textbf{\ours} & \textbf{0.319} & 0.314 & \textbf{0.371} & \textbf{0.504}\\
        \bottomrule
    \end{tabular}
    \label{tab:purify}
\end{table}

\begin{figure}[h]
    \centering
    \resizebox{\linewidth}{!}{
    \begin{tabular}{ccccccc}
        & \multicolumn{6}{c}{{\textbf{Editing Prompt}: `\textit{Set the background in a library}'.}}\\
        \rotatebox{90}{\footnotesize{\makebox[0.14\linewidth][c]{w/o Purification}}}
        & \hspace*{-2mm}\includegraphics[width=0.14\linewidth]{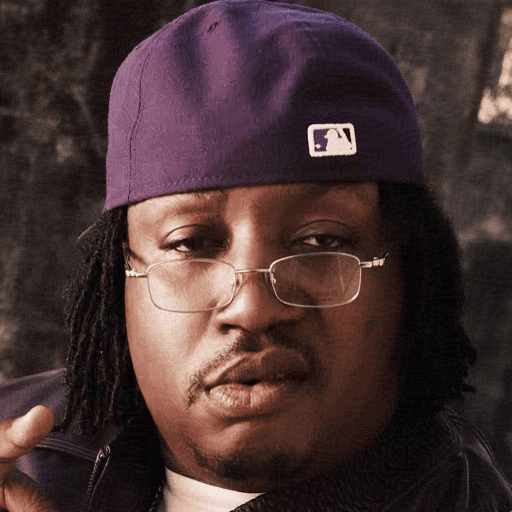} 
        & \hspace*{-3mm}\adjustbox{cframe=green}{\includegraphics[width=0.14\linewidth]{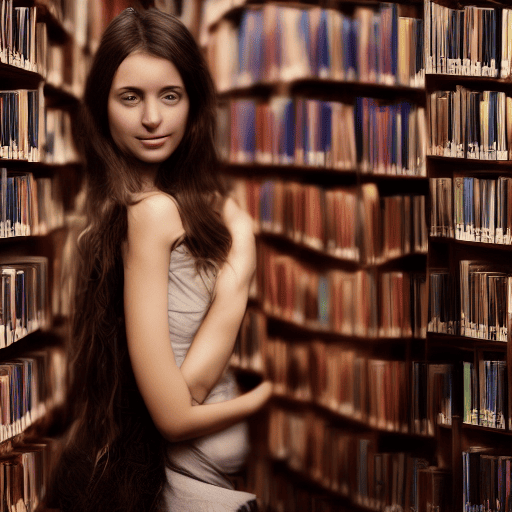}} 
        & \hspace*{-3mm}\includegraphics[width=0.14\linewidth]{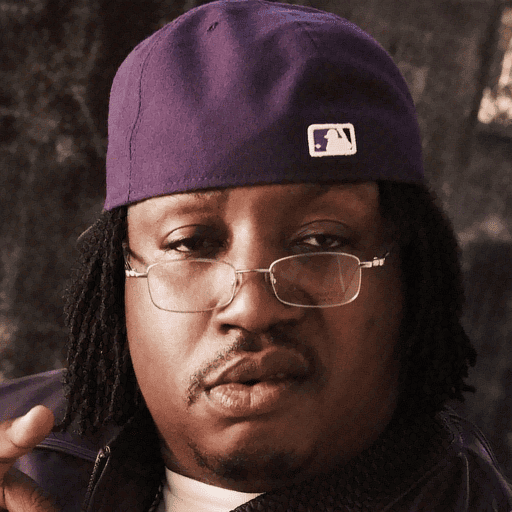} 
        & \hspace*{-3mm}\adjustbox{cframe=green}{\includegraphics[width=0.14\linewidth]{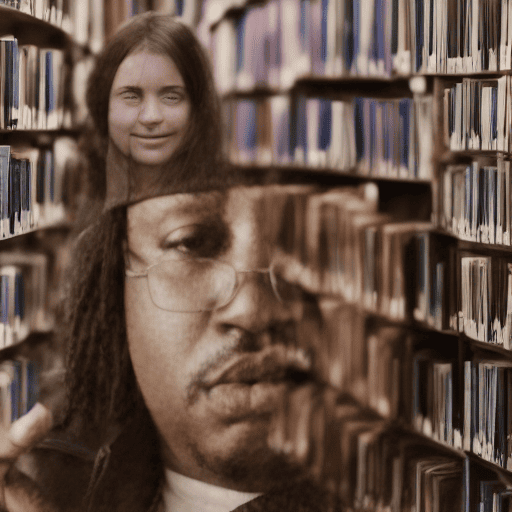}} 
        & \hspace*{-3mm}\includegraphics[width=0.14\linewidth]{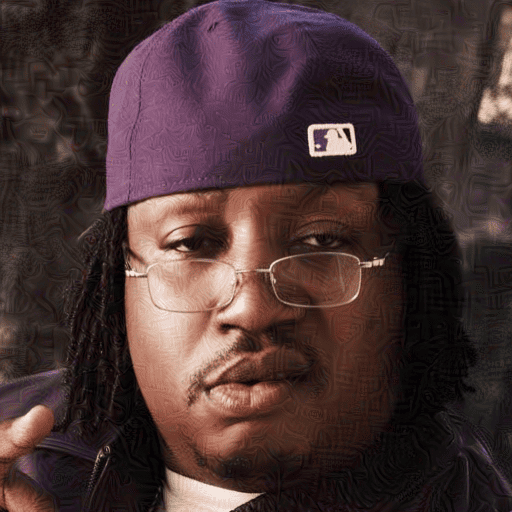} 
        & \hspace*{-3mm}\adjustbox{cframe=green}{\includegraphics[width=0.14\linewidth]{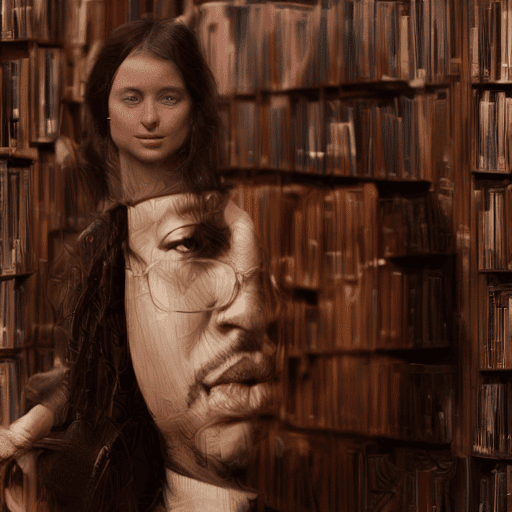}}\\
        \rotatebox{90}{\footnotesize{\makebox[0.14\linewidth][c]{Color Jitter}}}
        & \hspace*{-2mm}\includegraphics[width=0.14\linewidth]{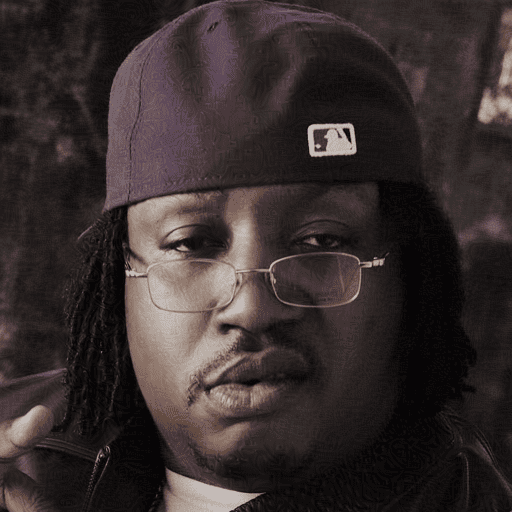} 
        & \hspace*{-3mm}\adjustbox{cframe=green}{\includegraphics[width=0.14\linewidth]{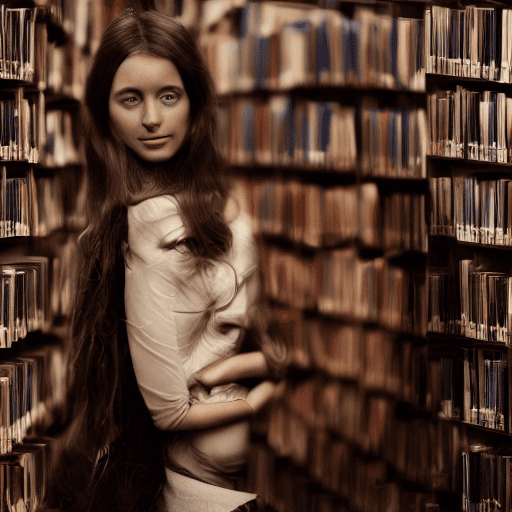}} 
        & \hspace*{-3mm}\includegraphics[width=0.14\linewidth]{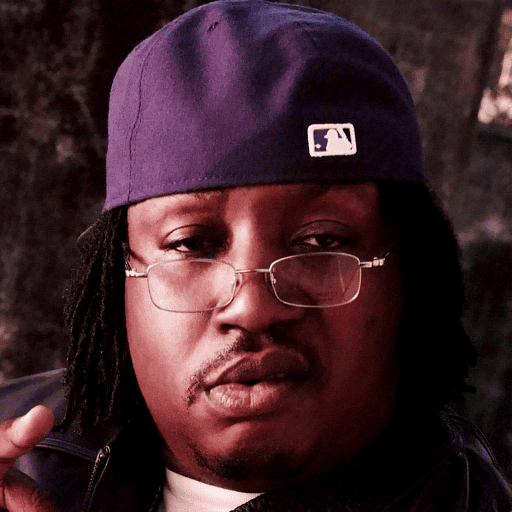} 
        & \hspace*{-3mm}{\includegraphics[width=0.14\linewidth]{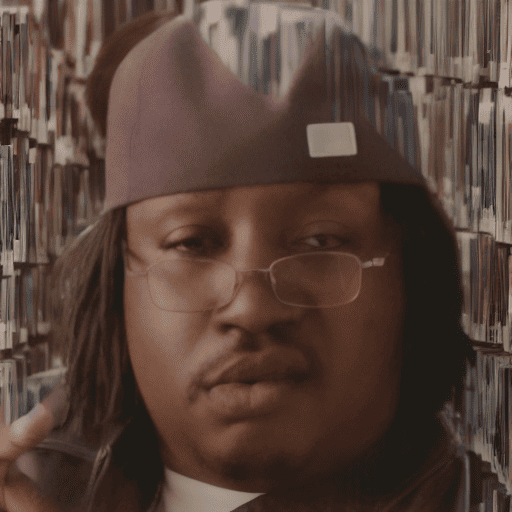}} 
        & \hspace*{-3mm}\includegraphics[width=0.14\linewidth]{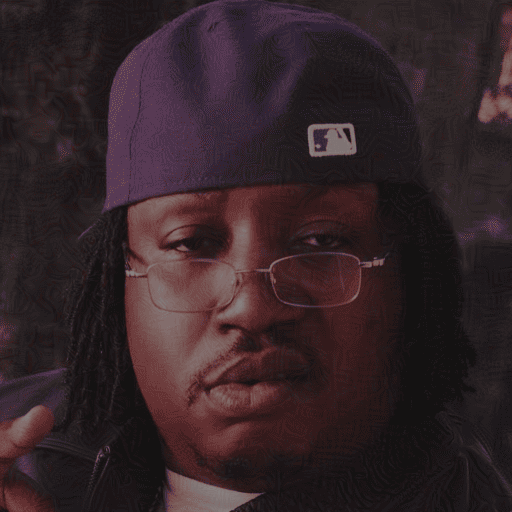} 
        & \hspace*{-3mm}\adjustbox{cframe=green}{\includegraphics[width=0.14\linewidth]{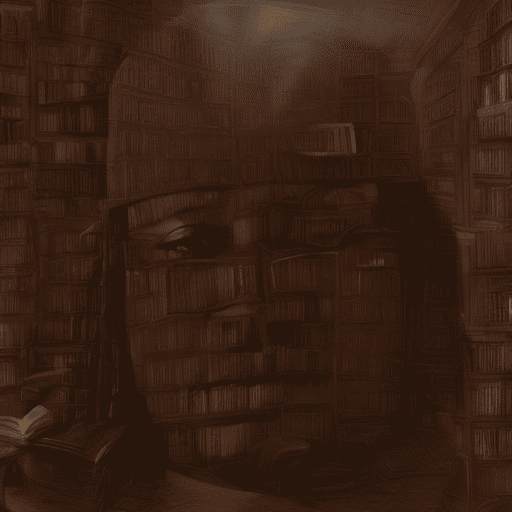}}\\
        \rotatebox{90}{\footnotesize{\makebox[0.14\linewidth][c]{DiffPure}}}
        & \hspace*{-2mm}\includegraphics[width=0.14\linewidth]{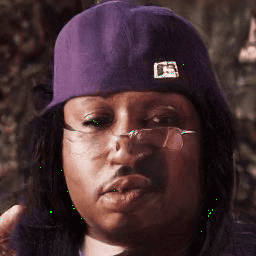} 
        & \hspace*{-3mm}\adjustbox{cframe=green}{\includegraphics[width=0.14\linewidth]{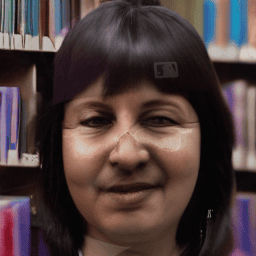}} 
        & \hspace*{-3mm}\includegraphics[width=0.14\linewidth]{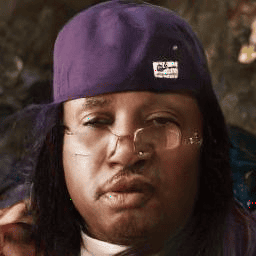}
        & \hspace*{-3mm}\includegraphics[width=0.14\linewidth]{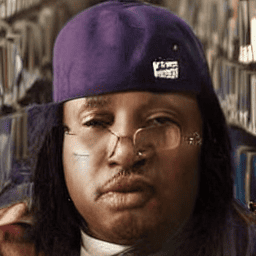}
        & \hspace*{-3mm}\includegraphics[width=0.14\linewidth]{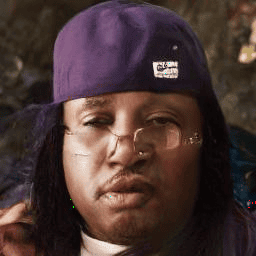} 
        & \hspace*{-3mm}\includegraphics[width=0.14\linewidth]{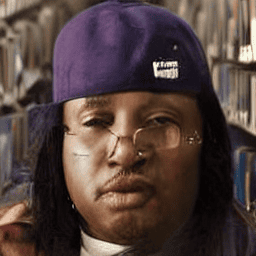}\\
        & \multicolumn{2}{c}{\hspace*{-3mm}\footnotesize{(a) \ours}}
        & \multicolumn{2}{c}{\hspace*{-3mm}\footnotesize{(b) PhotoGuard}}
        & \multicolumn{2}{c}{\hspace*{-3mm}\footnotesize{(c) EditShield}}\\
    \end{tabular}
    }
    \vspace{-0.8em}
    \caption{Qualitative results of edits on protected images after applying other purification methods. Images in green frames denote successful defense.}
    \label{fig:purify}
\end{figure}

To further assess the effectiveness of \ours, we evaluate its robustness against other purification techniques, namely Color Jitter and DiffPure~\cite{nie2022diffusion}. As presented in \textbf{Tab.~\ref{tab:purify}}, \ours consistently achieves the lowest FR scores (0.371 and 0.504) across both purification methods, demonstrating its ability to disrupt identity features after purification. While prior methods primarily interfere with edits, they fail to prevent identity retention post-purification. In contrast, \ours~ensures stronger identity removal while maintaining competitive LPIPS values, reinforcing its effectiveness as a defense mechanism. Qualitative results in \textbf{Fig.~\ref{fig:purify}} further supports these findings, showing that \ours~more effectively prevents identity recovery after purification.

\subsection{Impact of the FR Model}
To analyze the impact of the FR model, we conduct an ablation study comparing protection strength and efficiency with and without it. As shown in \textbf{Tab.~\ref{tab:fr_study}}, incorporating the FR model reduces the FR score from 0.534 to 0.316, achieving over 40\% improvement in identity protection. However, this comes with a slight increase in processing time per image (16s → 20s). Despite the added computational cost, these results highlight the necessity of the FR model for ensuring stronger identity protection.

\begin{table}[h]
    \centering
    \small
    \caption{Quantitative results on the effectiveness and efficiency impact of the FR model.}
    \vspace{-0.8em}
    \begin{tabular}{l|cc}
    \toprule
        Setting & Time/Image & FR ($\downarrow$) \\
    \midrule
         w/o FR Model & $\sim$16s & 0.534 \\
         w/ FR Model  & $\sim$20s & 0.316 \\
    \bottomrule
    \end{tabular}
    \label{tab:fr_study}
\end{table}


\end{document}